\documentclass[]{interact}

\usepackage{natbib}
\bibpunct[, ]{(}{)}{;}{a}{}{,}

\usepackage{graphicx}
\usepackage{siunitx}
\usepackage{tabularray}
\UseTblrLibrary{siunitx}

\usepackage{lineno}

\usepackage{pifont}
\usepackage{amsmath}
\usepackage{float}
\usepackage{subcaption}
\usepackage{enumitem}
\setlist[enumerate,2]{itemindent=0pt, leftmargin=24pt, label=(\alph*)}

\usepackage{array,multirow} 

\makeatletter
\newcommand{\captshort}{\let\blx@imc@ifciteseen\@firstoftwo}
\makeatother
\makeatletter
\newcommand*{\ForceNextCiteShort}{\AtNextCite{\let\ifciteseen\@firstoftwo}}
\newcommand*{\ForceNextCiteLong}{\AtNextCite{\let\ifciteseen\@secondoftwo}}
\makeatother

\usepackage[ruled, vlined, linesnumbered]{algorithm2e}

\SetAlCapHSkip{0em}  
\SetKw{Input}{Input: }
\SetKw{Output}{Output: }
\SetKw{Parameters}{Parameters: }
\SetKwProg{Fn}{}{:}{end}
\SetKwFor{While}{While}{:}{}
\SetKwFor{For}{For}{:}{}

\makeatletter
\patchcmd{\@algocf@start}
  {-1.5em}
  {0pt}
  {}{}
\makeatother

\newcommand{\geneticalgorithm}{genetic algorithm\xspace}

\newcommand{\genetic}{genetic\xspace}
\newcommand{\Genetic}{Genetic\xspace}

\NewDocumentCommand{\colorrule}{O{.4pt}m}{%
  {\color{#2}\hrule height#1}%
}

\usepackage{tabularx}
\newcolumntype{L}{>{\raggedright\arraybackslash}X}
\newcolumntype{Y}{>{\centering\arraybackslash}X}

\newcolumntype{t}{>{\hsize=.3\hsize}X}
\newcolumntype{s}{>{\hsize=.5\hsize}X}
\newcolumntype{b}{>{\hsize=1.5\hsize}X}

\usepackage[table,xcdraw]{xcolor}
\usepackage{colortbl}
\usepackage{tikz}
\usepackage{booktabs}

\usepackage{breakcites} 

\usepackage{amsthm}
\theoremstyle{remark}

\begin{document}


\title{Simulation-based optimization of a production system topology -\\ a neural network-assisted genetic algorithm}

\author{
N. Paape, J.A.W.M. van Eekelen, M.A. Reniers\thanks{Corresponding author: M.A. Reniers. Email: m.a.reniers@tue.nl}\\
\affil{Eindhoven University of Technology, Eindhoven, The Netherlands}
}

\maketitle
\begin{abstract}
There is an abundance of prior research on the optimization of production systems, but there is a research gap when it comes to optimizing which components should be included in a design, and how they should be connected. To overcome this gap, a novel approach is presented for topology optimization of production systems using a genetic algorithm (GA). This GA employs similarity-based mutation and recombination for the creation of offspring, and discrete-event simulation for fitness evaluation. To reduce computational cost, an extension to the GA is presented in which a neural network functions as a surrogate model for simulation. Three types of neural networks are compared, and the type most effective as a surrogate model is chosen based on its optimization performance and computational cost.

Both the unassisted GA and neural network-assisted GA are applied to an industrial case study and a scalability case study. These show that both approaches are effective at finding the optimal solution in industrial settings, and both scale well as the number of potential solutions increases, with the neural network-assisted GA having the better scalability of the two. 
\end{abstract}

\section{Introduction}\label{sec:introduction}
Production systems require sizeable investments to develop, but their performance is often difficult to predict, let alone guarantee upfront. As such, simulation has become indispensable in the production system design process, especially for selecting the optimal design from its alternatives. One of the challenges in using simulation effectively is computational cost \citep{Horng2021}.

Modern production systems feature complex components and dynamics, leading to larger, computationally expensive simulation models \citep{Wuest2016}.  Additionally, there is an increasing need for flexible systems that perform well across diverse conditions \citep{Florescu2020}. As a result, extensive simulation of various production scenarios is required. Furthermore, the design process of these systems involves a multitude of interconnected design choices, with each design parameter exponentially increasing the number of viable designs \citep{Gunn2022}. 
Altogether, this means that fully exploring the design space of a system involves extensive simulation for numerous designs with computationally expensive models.

This work focuses on optimizing the topological design of a production system, that describes which components are featured in the system and how they are connected to each other (an example is shown in Figure~\ref{fig:design_space_example}, which is explained in Section~\ref{sec:design}). For the reasons stated above, fully exploring the design space of topologies for a complex system can be infeasible. To address this issue, a genetic algorithm is proposed that helps to identify the best topology while simulating only a fraction of all possible designs. To further reduce computational cost, the genetic algorithm is extended with a neural network that functions as a cheap-to-evaluate surrogate model for simulation.


\subsection{State-of-the-art}\label{subsec:state-of-the-art}
This section reviews the state-of-the-art on simulation in the design of production systems, on simulation-based optimization of production systems, on the use of surrogate models in optimization, and on the optimization of topologies.

\subsubsection{Simulation in the design of production systems}\label{subsubsec:simulation_production_systems}
Considerable research is dedicated to (discrete-event) simulation in production system design, with recent overviews given in \cite{Mourtzis2020} and \cite{DePaulaFerreira2020}. One of the most used methods for analyzing product flow in production systems is discrete-event simulation (DES) \citep{GalvaoScheidegger2018}. DES models are akin to expensive black-box functions \citep{Zmijewski2020}; their computational cost is high \citep{Rabe2020}, and there generally is no simple mathematical model describing the input-output relation \citep{Gunn2022}.  

The traditional simulation-based design process is an iterative trial-and-error process of system design, model creation, simulation, and re-design based on the predicted performance \citep{Mollaghasemi1998}.
However, the trial-and-error process, coupled with frequent (re)modeling and time-consuming simulation, is ineffective for problems with large design spaces \citep{Sabuncuoglu2002}, and for problems where there is a complex relationship between topology and performance. More recent approaches such as \cite{Gunn2022} use simulation-based optimization to automate this process, but these approaches generally rely on a parametrization of the design space, and it is not immediately clear on how these methods can be used to optimize a production system's topology.


In \cite{Paape2023c}, the authors of this article introduced a method for automated \textit{design space exploration} for production system topologies. This approach involves automatically exploring and analyzing the \textit{design space} -- the set of all feasible designs -- to identify the most promising designs. This is achieved by automatically iterating through the design space, constructing DES models, simulating production scenarios, and evaluating the chosen performance indicators. One of the bottlenecks identified when applying this method to a real-world case study was in dealing with the computational cost of simulating all feasible designs, which for complex systems is often insurmountable. As such, more efficient optimization strategies are required for automated design space exploration to be effective.

\subsubsection{Simulation-based optimization of production systems}
There is extensive prior research on simulation-based optimization of production systems. A survey on simulation-based optimization in the context of manufacturing is presented in \cite{Liu2018a}. Although the survey focuses on manufacturing operations and not design, it provides a good overview of optimization strategies. 

According to \cite{Liu2018a}, the most frequently utilized global optimization strategies in manufacturing are evolutionary algorithms (EA), simulated annealing (SA), and tabu search (TS), which all have the advantage of being able to deal with highly-dimensional problems with both discrete and continuous decision spaces. In contrast, Bayesian optimization, another method widely employed in manufacturing \citep{Zmijewski2020, Gunn2022}, is better suited for optimization problems with lower dimensionality and continuous variables \citep{Frazier2018}. \citeauthor{Liu2018a} concludes that of EA, SA, and TS, the method that has been utilized the most for complex manufacturing systems is EA, with the most well-known variant being the genetic algorithm. In addition to being able to deal with discrete and continuous numerical decision spaces, some variants of evolutionary algorithms also support other types of decision spaces. For example, \cite{Dushatskiy2019} deals with categorical decision spaces, which means that variables represent categories or groups, and can only have a finite number of possible values. \cite{Li2016} deals with topological decision spaces, meaning that decisions made for a design are on which nodes or edges are included.

\subsubsection{The use of surrogate models in optimization}
Evolutionary algorithms are often used in combination with efficiency-increasing techniques to help cope with the high computational complexity of (discrete-event) simulation \citep{Liu2018a, Hoad2015}
. A frequently used approach in simulation-based optimization is to use surrogate models (also referred to as auxiliary models or metamodels). A surrogate model is an approximation of the simulation model, that is computationally cheaper to evaluate, thus facilitating faster optimization. A recent literature review is given in \cite{SoaresdoAmaral2022}. The types of surrogate models most often encountered in literature are Kriging, Polynomial regression, and artificial neural networks (ANN). As stated in \cite{Barkanyi2021}, polynomial regression is effective when underlying models are of low complexity. Kriging is most suitable for capturing smooth functions with low dimensionality ($<20$) and with continuous variables. In contrast, ANNs are effective for capturing the behavior of high-dimensional nonlinear systems \citep{Barkanyi2021, Wuest2016}. This makes ANN suitable for production system optimization which often deals with highly dimensional and discrete parameter spaces.

Combining simulation and neural networks in the design process of production systems has been a long-standing concept \citep{Mollaghasemi1998}. An example of simulation-based optimization using neural networks is shown in \cite{Vosniakos2006}, where a DES model of a manufacturing cell is used to train a neural network. A genetic algorithm then optimizes the manufacturing cell design, employing the neural network as a surrogate model to speed up computation.
%
%
In \cite{Azimi2017}, an optimization method for the facility layout problem is introduced. A discrete-event simulation model is employed to generate data on the makespan of layouts, which is then used to train an artificial neural network. This ANN serves as a surrogate model in a non-dominated sorting genetic algorithm (NSGA-II). The method is validated using a real case study in manufacturing with hundreds of millions of potential designs. This study showed that a significant computational cost reduction can be achieved when optimizing large design spaces.


Similar approaches are also found for other simulation types. In the petroleum industry, neural networks have been used as surrogate models for numerical simulation in genetic algorithm-based optimization \citep{Golzari2015}. 
\cite{Pfrommer2018} presents a method for the optimization of manufacturing processes using Differential Evolution (DE) and a deep neural network which serves as a surrogate model for Finite Element simulation. Finally, in \cite{Esche2022}, neural networks are used as surrogate models for dynamic systems models in chemical engineering applications.

There are also plenty of examples of other viable types of surrogate models. \cite{Horng2021} proposes a method for computationally expensive simulation optimization problems. The method uses Multivariate Adaptive Regression Splines (MARS) as a surrogate for simulation, in combination with ordinal optimization and ant lion optimization. The proposed method is demonstrated using a communication system optimization problem. 
\cite{Gunn2022} describes a method combining discrete-event simulation with Bayesian optimization. This method is effectively utilized in a carbon fiber wheel manufacturing case study, in which input parameters such as the cycle times of machines, and number of tools are optimized to maximize throughput. 
In \cite{DeUgarte2011} a genetic algorithm is used in the online optimization of a production schedule. The quality of a solution is evaluated using discrete-event simulation. The genetic algorithm uses evaluation relaxation to reduce computation time, which is achieved through a predator mechanism that terminates simulation prematurely if the solution is deemed of poor quality.



\subsubsection{Topology optimization}
Previously mentioned optimization methods aim to solve optimization problems with a numerical decision space, be it continuous and/or discrete. However, the focus of this paper is on optimizing the topology of a production system, for which prior work is scarce.
An area of research that deals with similar topological design problems is molecular chemistry, where the goal is to find new molecules with desired properties. In \cite{Kwon2021a} an evolutionary algorithm is proposed for optimizing new molecules with desired properties. In the proposed method, a chemical library is used to train a neural network, which acts as a model for fitness evaluation. One of the main differences with simulation-based optimization is that there is no known function to evaluate the fitness of a solution. 
\cite{Li2016} proposes a multi-objective evolutionary algorithm for optimizing the topology of a communication network. In this work, no surrogate model is required, as the calculation of the objective function is numerical and relatively inexpensive. 
In \cite{Dushatskiy2019}, a surrogate-assisted GA is proposed for expensive-to-evaluate optimization problems with (binary) categorical variables. The author suggests that this algorithm can be used to optimize architectural choices of deep neural networks. The GA utilizes a convolutional neural network as a surrogate model, in combination with pairwise regression, meaning it is trained to identify the difference between pairs of solutions. The advantage of pairwise regression is that fewer evaluations are needed to train the neural network. 

\subsection{Contribution}
Although simulation-based optimization has been widely applied in manufacturing, there does not yet seem to be any prior work on  topology optimization for production systems. There is some prior work on topology optimization in other fields of research, but these are often very different in nature. For example, in \cite{Li2016} the computational cost of evaluating the objective function is significantly lower than that of a discrete-event simulation model. In \cite{Dushatskiy2019}, it is hypothesized that the developed algorithm could work for the optimization of architectural choices in deep neural networks. However, it is unclear whether this approach would work for topology optimization of production systems. Finally, the method developed in the work of \cite{Kwon2021a} requires that a topology is first encoded into a datatype specific for molecular chemistry, making it unsuitable for manufacturing applications.

The goal of this work is to establish a method for topology optimization for production systems. This method requires efficiency-increasing techniques to deal with computational cost. This leads to the main contributions of this work: 
\begin{itemize}
    \item A \geneticalgorithm for topological optimization of production systems that uses similarity-based mutation and recombination for offspring creation, and discrete-event simulation for fitness evaluation. 
    \item An extension to this algorithm which incorporates a neural network as a surrogate for discrete-event simulation. 
\end{itemize}


Table~\ref{tab:state-of-the-art} shows how this contribution compares to the state-of-the-art. Besides the main contributions, the following sub-contributions are made: 
\begin{itemize}
    \item  A feedforward ANN, pairwise ANN, and Bayesian ANN architecture are compared to determine which is most effective as a surrogate model.
    \item The effectiveness of the algorithms (with and without a surrogate model) for use in industrial applications is evaluated using a case study in poultry processing.
    \item The scalability of the optimization algorithm (with and without a surrogate model) is evaluated using a benchmark case study of a loop layout system. 
\end{itemize}

\begin{table}[hpbt]
\caption{This work is compared to the state-of-the-art. Below the table, a glossary for the abbreviations is given.}
\label{tab:state-of-the-art}
\centering
\footnotesize
\setlength{\tabcolsep}{4pt}
\begin{tabularx}{\linewidth}{ | 
>{\raggedright\hsize=0.18\hsize}X |   
>{\centering\hsize=0.1\hsize}X |    
>{\centering\hsize=0.1\hsize}X |    
>{\centering\hsize=0.2\hsize}X |    
>{\centering\hsize=0.13\hsize}X |    
>{\centering\hsize=0.07\hsize}X |    
>{\centering\arraybackslash\hsize=0.22\hsize}X|     
}     
\hline
 & \textbf{Optimi-zation} & \textbf{Model} & \textbf{Surrogate} & \textbf{Decision space} & \textbf{Obje-ctive} & \textbf{Application domain}\\ \hline 
\textbf{\citealt{Vosniakos2006}} & GA  & DES & Feedforward ANN & Discrete \linebreak numerical & Single & Manufacturing cells \linebreak (Manufacturing)  \\ \hline
\textbf{\citealt{DeUgarte2011}} & GA  & DES & Evaluation Relaxation & Categorical & Single& Scheduling \linebreak (Manufacturing)   \\ \hline
\textbf{\citealt{Golzari2015}} & GA  & NS & Feedforward ANN & Continuous \linebreak numerical & Single & Production control of petroleum wells \\ \hline
\textbf{\citealt{Li2016}} & NSGA-II \linebreak  & Analytic & None & Topological & Multi & Communication networks \\ \hline
\textbf{\citealt{Azimi2017}} & NSGA-II \linebreak  & DES & Feedforward ANN & Categorical & Multi & Facility layout \linebreak (Manufacturing)  \\ \hline
\textbf{\citealt{Pfrommer2018}} & DE  & Finite Element & Deep ANN & Continuous \linebreak numerical & Single & Composite textile \linebreak draping process \linebreak (Manufacturing)  \\ \hline
\textbf{\citealt{Dushatskiy2019}} & GA  & Generic & Convolutional Pairwise ANN & Categorical \linebreak (binary) & Single & Generic \\ \hline
\textbf{\citealt{Zmijewski2020}} & BO & DES &  Gaussian Process & Discrete \linebreak numerical & Single & Resource allocation \linebreak (Manufacturing)  \\ \hline
\textbf{\citealt{Kwon2021a}} & GA  & None$^{*}$ & Deep ANN & Topological & Single & Molecular chemistry \\ \hline
\textbf{\citealt{Horng2021}} & OALO & Generic & MARS & Numerical & Single & Generic \\ \hline
\textbf{\citealt{Gunn2022}} & BO & DES & Gaussian Process & Continuous \linebreak \& discrete \linebreak  numerical & Single & Production process optimization \linebreak (Manufacturing)  \\ \hline
\textbf{This work} & GA  & DES & Feedforward ANN, \linebreak Pairwise ANN, \linebreak or Bayesian ANN
& Topological$^{\dagger}$ & Single & Production system topology opt. (Manufacturing) \\ \hline

\end{tabularx}
\vspace{1mm}
\linebreak
\raggedright
*: The surrogate model is built using a database.\\
$\dagger$: In this work, the topologies are first converted to binary categorical data. \\
\vspace{2mm}
\hrule
\vspace{2mm}
\textbf{Abbreviations}\\
BO: Bayesian Optimization\\
GA: Genetic Algorithm\\
NSGA-II: Non-dominated Sorting Genetic Algorithm II\\
DE: Differential Evolution\\
OALO: Ordinal Optimization and Ant Lion Optimization\\
DES: Discrete-Event Simulation\\
NS: Numerical Simulation\\
ANN: Artificial Neural Network\\
MARS: Multivariate Adaptive Regression Splines\\
\hrule

\end{table}

\subsection{Structure of the paper}
This work is organized as follows. Section~\ref{sec:design} describes the definition of a `design' in the context of production system topology. In Section~\ref{sec:case_studies} the industrial case study and scalability case study are described. Section~\ref{sec:ga} outlines the \geneticalgorithm for optimization of production system topologies and its performance when applied to the case studies.
The extension to the algorithm with a neural network surrogate model is discussed in Section~\ref{sec:nn-ga}, along with results for the case studies. Finally, concluding remarks are given in Section~\ref{sec:conclusion}.


\section{Production system topology}\label{sec:design}
\begin{figure}[phtb]
    \centering
    \includegraphics[width=0.87\linewidth]{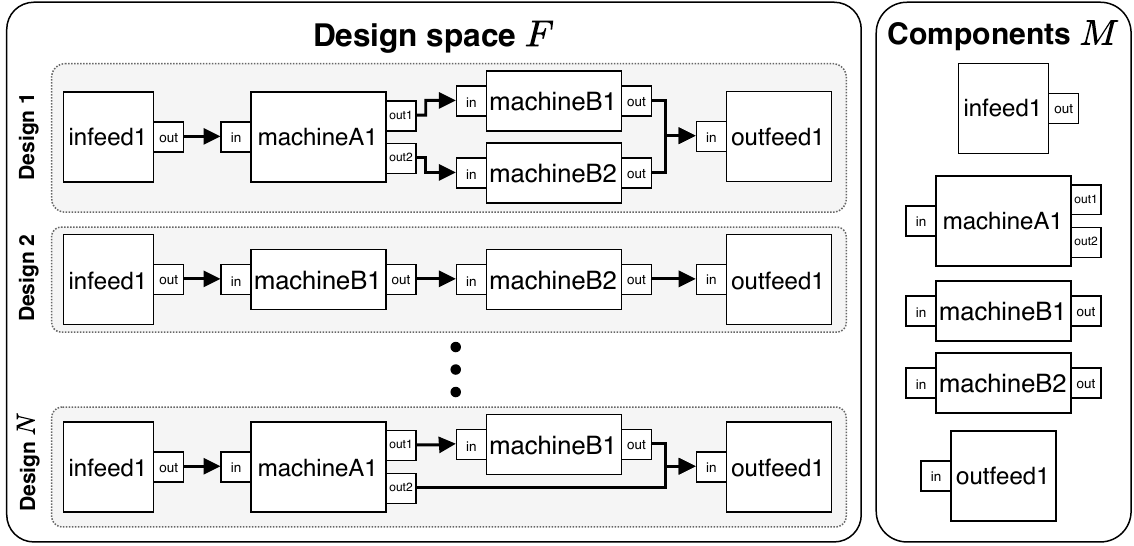}
    \caption{Example of a design space $F$ with $N$ feasible designs. The set of component instances $M$ contains the five possible component instances from four different component types (Infeed, MachineA, MachineB, and Outfeed).}
    \label{fig:design_space_example}
\end{figure}

This section specifies what a design and design space are in the context of a production system topology. 
In this work, a \textit{design} refers to the topology of a production system, which describes the components of the system and the connections between them. Examples of designs are shown in Figure~\ref{fig:design_space_example}. In this context, the topology is regarded as a graph, with the nodes and edges representing the components and connections in the system. Each node is an instance of a component type, with each component type having (multiple) input and/or output ports. The edges describing how components are connected are unidirectional from output port to input port. These types of graphs are referred to as labeled port graphs \citep{Fernandez2018}.

The design space is the set of feasible designs of the system, which is denoted by $F$. In Figure~\ref{fig:design_space_example} an example of a design space with $N$ designs is given. In this design space, there are a maximum of $5$ different components that can be connected in various ways. It is assumed that the set of all feasible designs is known. In previous work \citep{Paape2023}, the authors of this paper propose a method for automatically generating this set using a syntax-based specification language.



Each feasible design contains a finite number of components and, more often than not, each component will feature in multiple designs. $M$ is the set of all possible components that are featured in all designs of the design space. It is the set of all nodes that can potentially be present in a feasible topology. $O$ and $I$ are respectively the sets of all output ports and input ports of components in $M$.




\section{Case studies}\label{sec:case_studies}
In this work, two case studies are analyzed. The first is based on the poultry processing system case study presented in \cite{Paape2023c}, and is used to analyze optimization performance in an industrial setting. The second is a study based on the loop layout benchmark problem in \citep{Saravanan2015}, and is used to analyze scalability. For each of the case studies, all possible designs were simulated. This makes it possible to determine whether an optimization algorithm finds the optimal design. Simulating all designs is not necessary in practical applications.

\subsection{Poultry processing case study}\label{sec:poultry_case}
\begin{figure}[bthp]
    \centering
    \includegraphics[width=1\linewidth]{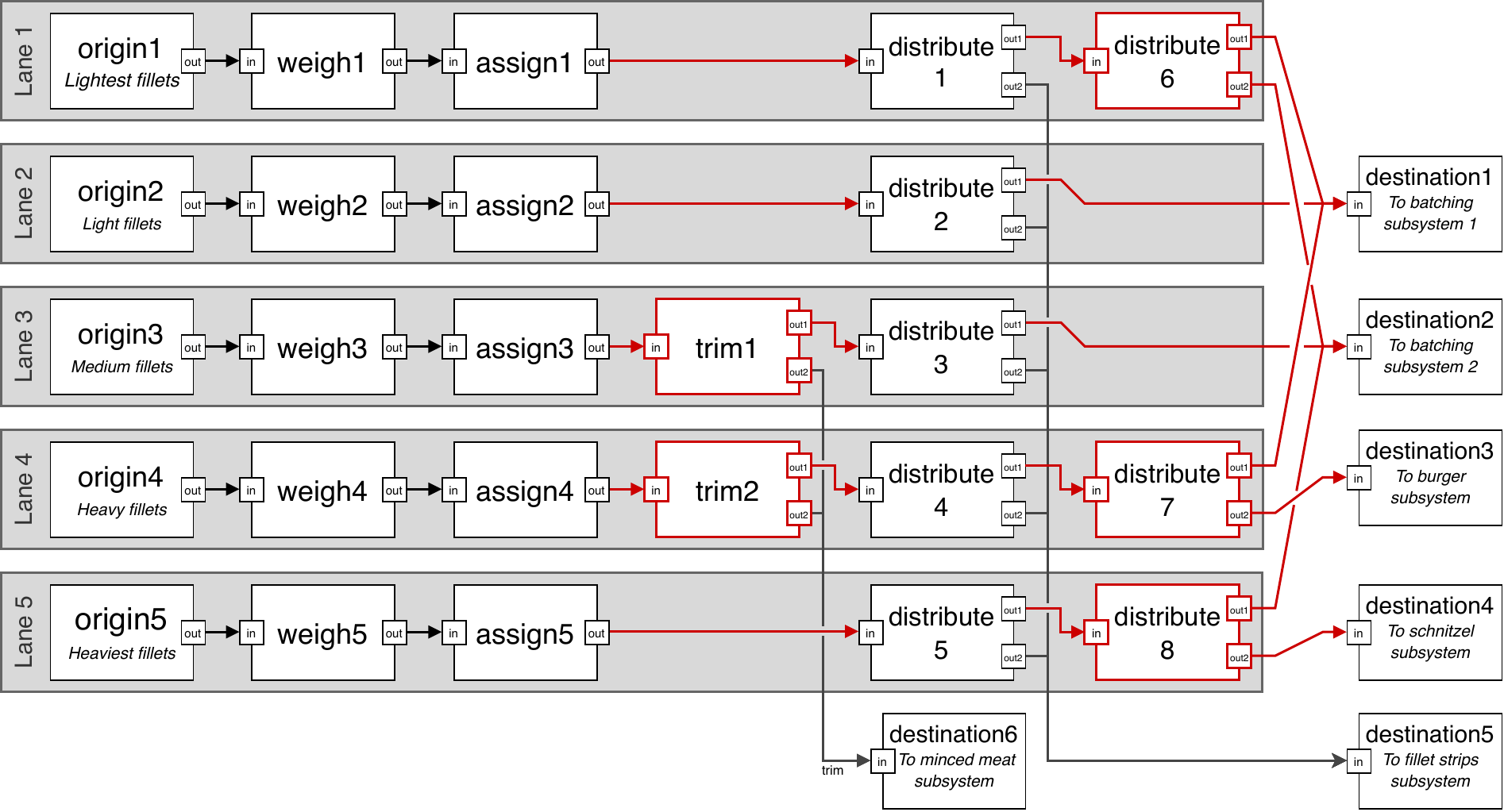}
    \caption{The poultry fillet distribution system from \cite{Paape2023c}. In red are component instances and connections which can differ per design. Some components and connections are not present in this specific design. For example, in an alternative design `assign5' might instead be connected to a `trim3' component.}
    \label{fig:case_poultry}
\end{figure}

The first case study is based on the poultry fillet distribution system described in \cite{Paape2023c}. A more in-depth description of the case study can be found there. The purpose of this system is to distribute fillets of different weights arriving from upstream systems, to the most suitable downstream destinations, with each destination having specific weight requirements. The goal is to design a system that maximizes the number of fillets that arrive at the correct destinations, while minimizing the cost of the design. Figure~\ref{fig:case_poultry} shows a possible system topology. 

The system consists of five lanes, each dealing with its class of fillet weights ranging from light to heavy. 
Each lane starts with an entry point, followed by a weighing module, and a module where fillets are assigned to a destination
. Next is an optional trim module, where fillets can be trimmed down to make it more suitable for the chosen destination. The trimmed parts are sent to `destination6'. The last step in this system is distribution, in which fillets are distributed to the assigned destinations. Each lane has one distributor that can route otherwise unsuitable fillets to `destination5', and potentially one additional distributor for the other destinations.

A small-scale and large-scale design problem are analyzed. The first is identical to the design problem specified in \cite{Paape2023c}, which has 11520 possible topologies. For this experiment, a design can have three distributors that can be placed freely in addition to the five default distributors. Each lane optionally has a trimming module, but each additional module comes at a penalty in the objective function. The modules and connections that are specific to the design in Figure~\ref{fig:case_poultry} are highlighted in red. 

The large-scale experiment is similar, but has five distributors that can be placed freely instead of three. This change results in a total of 190464 possible topologies. The goal is to analyze optimization performance for an industrial case study with a larger design space. To reduce the computation time required for simulating all designs, each design was simulated for a reduced duration, and for fewer production scenarios (compared to the case study in the original paper).

\subsection{Scalability case study}\label{sec:benchmark_case}
As noted in \cite{Liu2018a}, one of the key issues is the lack of benchmark problems for simulation-based optimization in manufacturing. To the knowledge of the authors, there is no benchmark for topology optimization of production systems. Therefore, the benchmark found in \cite{Saravanan2015} was modified to make it suitable for the intent of this work. This benchmark is based on the loop layout problem in flexible manufacturing systems to analyze the scalability of the algorithms. 

\begin{figure}[t]
    \centering
    \includegraphics[width=.75\linewidth]{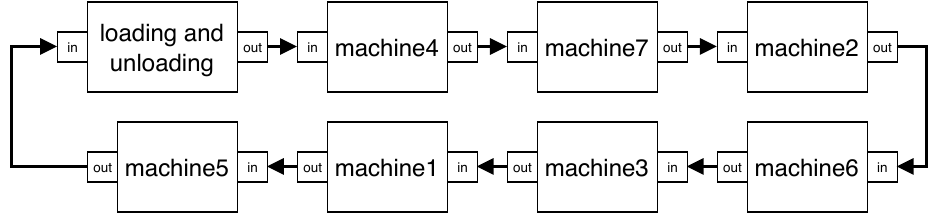}
    \caption{A potential 7-machine loop layout, based on \cite{Saravanan2015}.}
    \label{fig:case_benchmark}
\end{figure}

The case study is as follows. The production system is built up out of $n$ machine stations and one station for loading and unloading. These are configured in a unidirectional loop with parts flowing from station to station; when going around the loop each station is visited. In Figure~\ref{fig:case_benchmark} a possible topology of a 7-machine loop layout is shown. Parts enter the system at the loading/unloading station and exit once processing has been completed.  Each part has a specific processing plan; the sequence of machines it needs to visit to be processed. This processing plan must be followed in the specified order, and might require multiple laps around the system. Thus, the order in which machines are placed has a big effect on the amount of time that parts spend in the system. The goal is to find the optimal system topology that minimizes this time.

The following changes were made to make the study more suitable for topology optimization. The first change is in how designs are represented. In \cite{Saravanan2015} a design is considered a sequence of machines. In this work, a design is instead considered as a topology. The second change is in the complexity of the model. The original benchmark study has a straightforward objective function that is calculated analytically.
The modified case study introduces more complex dynamics and requires discrete-event simulation. The modifications were the incorporation of inter-arrival times for parts ($4$ seconds), processing times for machines ($5$ seconds), and transportation times for moving from station to station ($1$ second). Additionally, each machine now has a finite buffer (of size $2$). If a part arrives when the buffer is full, the part continues on to the next machine and will have to go around the loop once more. 
The number of parts that can be in the loop at once is not constrained. 

A $n$-machine system produces one `family' of parts, for which there is one `blueprint', which is a sequence of operations of length $n$. The different `processing plans' that the system produces are all possible sub-sequences of this blueprint. Each part processed by the system receives a processing plan at random. E.g., for a 3-machine loop for which the blueprint is [1-2-3], the allowed processing plans are: [1-2-3], [1-2], [2-3], [1], [2], and [3] (sequences such as [1-3-2] or [1-3] are not allowed). The goal is to identify the topology with the lowest mean cycle time for this family of parts. For each possible system topology a DES model is created and is simulated for a total of 2 hours of simulation time.

In total, four design problems are analyzed: a 6-, 7-, 8-, and 9-machine loop system. Each problem has $(n!)$ potential designs, resulting in, respectively, $720$, $5040$, $40320$, and $362880$ designs. The purpose of this study is to gain insight into the scalability of the algorithms presented in this paper.


\section{Genetic optimization}\label{sec:ga}
\Genetic algorithms use the principles of natural selection and evolution to find solutions to optimization problems. The premise behind a genetic algorithm is to iteratively refine a population of solutions over multiple generations. The four main concepts in a genetic algorithm are recombination, mutation, evaluation, and selection. \textit{Recombination} combines `genetic information' from selected parent solutions to produce offspring, \textit{mutation} introduces random changes to parent solutions to produce offspring, \textit{evaluation} involves assessing the fitness of new offspring, and \textit{selection} determines which solutions are chosen to be the next generation. The specific contribution of this section is in how these general concepts can be implemented with the goal of genetic optimization of production system topologies. The next sections explain the genetic representation of a design, the considerations and implementations of  evaluation, selection, mutation, and recombination, the resulting genetic algorithm and its parameters, and its performance in the case studies.\\



\subsection{Genetic encoding of a design}\label{sec:genetic_encoding}
\begin{figure}[hbpt]
    \centering
    \includegraphics[width=0.8\linewidth]{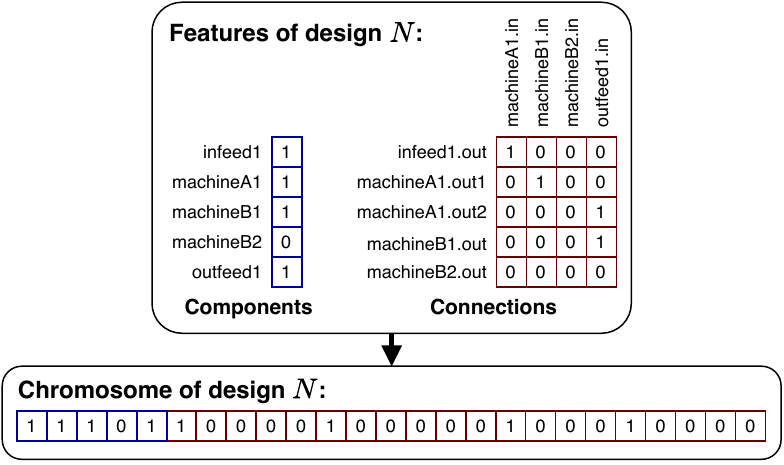}
    \caption{The chromosome representation of design $N$ as seen in Figure~\ref{fig:design_space_example}.}
    \label{fig:encoding_example}
\end{figure}

In genetic algorithms, solutions are typically encoded through so-called chromosomes, which are the genetic representation of a given solution. Chromosomes are built up using genes, with each gene usually describing the value of one of the parameters in the decision space. 
Representing a production system topology as a chromosome is not trivial, which is why the topologies are first converted to binary categorical data. The `nodes' of a topology are encoded as a binary vector of length $|M|$ in which each element represents whether a component $m \in M$ is present in the topology. The `edges' of the topology can be encoded as a connection matrix of size $|O|\times|I|$, in which the rows and columns represent all possible output and input ports in the system respectively, and each element has a boolean value that represents whether there is a connection from port $o \in O$ to port $i \in I$. Naturally, if the component corresponding to either port $o$ or $i$ is not included in the topology, the connection cannot be included either. Together the encoding of the nodes and edges make up the chromosome of a design. Figure~\ref{fig:encoding_example} shows the chromosome representation of design $N$ as seen in Figure~\ref{fig:design_space_example}. 


\subsection{Evaluation}
The first of these concepts which will be explained, is evaluation, the step in which the fitness of a solution is evaluated through discrete-event simulation. Using discrete-event simulation for topology optimization of production systems, requires that for each unique topology a new DES model is generated. DES models can be built modularly by instantiating model components and connecting their ports \citep{Zeigler1987}. Such a model is built up in a way that resembles the labelled port graphs which are used to describe the topology of a production system, making it possible to automatically generate the DES model for a given topology. 
%
%
This model is then simulated for the selected production scenarios, after which the fitness of the solution is determined using a chosen objective function. This assumes that models can be automatically generated and simulated based on the given topology and the specified production scenarios. This method is not viable if the model needs to be tailor-made to each design. This means this method is only viable if the production system and production controller can automatically be adjusted to the specific topology of the system. It does not work if a system designer needs to finalize the design manually, or if the production schedule for a system is hand-made by an operator. 

The individual model components must be carefully constructed and validated, as validating the models of each and every design is infeasible. Further complicating fitness evaluation is that production processes are often stochastic, which demands careful tuning of simulation parameters such as simulation duration, simulation warm-up, and number of independent simulation runs \citep{Law2015}. 



        

\subsection{Selection}
The selection step determines which solutions survive to the next population. For a genetic algorithm to work requires there to be evolutionary pressure: better-performing solutions must have a higher chance of surviving (exploitation). To allow for hill climbing, suboptimal solutions must have some chance of survival (exploration). An overview of selection strategies is given in \cite{Yadav2017}. The most frequently used approaches are elitism, roulette wheel, tournament selection, and ranked. 

In elitism, the best solutions are selected for exploitation, and a few random solutions are allowed for exploration. The main downsides are that its exploitation component lacks diversity, while its exploration component lacks evolutionary pressure.
In tournament selection, $n$ solutions are randomly picked for a tournament, with the winner advancing to the next generation. It combines exploitation (best solution wins) and exploration (randomly chosen solutions in each tournament). However, determining $n$ for a balanced exploration-exploitation trade-off is not straightforward.
In roulette wheel selection, the probability of being selected is proportional to a solution's fitness. A solution's fitness is the performance evaluated through simulation. Roulette wheel selection usually leads to a good balance between exploration and exploitation, although it has difficulties when fitness differences are extremely large or small.

\begin{algorithm}[tb]
\DontPrintSemicolon
\caption{The rank-based selection algorithm}\label{alg:selection}
\SetKwFunction{FnSelect}{Select}%
\Parameters{Selection pressure $\alpha_s$, population size $\beta$.}\;
\Input{A set of solutions $X$ and a fitness function $f$.}\;
\BlankLine
    Initialize the set of selected solutions: $Y \leftarrow \emptyset$.\;
    \While{desired population size is not reached $|Y| \leq \beta$, and there are possible solutions remaining $|X|>0$}{
        Determine the rank $r_x$ for each solution $x \in X$ based on fitness function $f(x)$.\;
        Calculate the probability for each solution $x$ to be selected: $P(x) = \frac{e^{\alpha_s \cdot (r_x - 1)} - 1}{e^{\alpha_s \cdot (|X| - 1)} - 1}$.\;
        Based on these probabilities, select a random solution $x$.\;
        Add chosen solution $x$ to $Y$, and remove it from $X$.\;
    }
\BlankLine
\Output{set of selected solutions $Y$.}\;
\end{algorithm}

In this work, exponential ranked selection is used. Ranked selection is similar to roulette wheel, but instead of probability scaling with fitness, it scales with solution rank. The rank of a solution is its ranking when ordering the solutions based on their fitness. For exponential ranking, the probability of a solution $x$ with rank $r_x$ being selected is $P(x) = \frac{e^{\alpha \cdot (r_x - 1)} - 1}{e^{\alpha \cdot (N - 1)} - 1}$, with number of solutions considered $N$, and evolutionary pressure $\alpha$. A lower $\alpha$ promotes more diverse solutions, resulting in slower convergence but a higher probability of finding the optimum, and vice versa. 

Algorithm~\ref{alg:selection} shows the algorithm for exponential rank-based selection used in this paper. The fitness function $f(x)$ is used to rank the solutions $x \in X$. When selecting which population survives to the next generation, the fitness is the performance of the design evaluated using discrete-event simulation. Later in this section, other steps of the algorithm will use the rank-based selection approach with other fitness functions.

\subsection{Mutation}
In mutation, random changes are introduced to the solutions of the current population, with the hope that some of these changes result in improved performance. The most simple form of mutation is to invert `bits' in a chromosome with a small probability \citep{Back1993}. Depending on the optimization problem, this might result in an infeasible solution. There are many ways to deal with infeasible solutions, for example, to discard infeasible solutions entirely, to penalize them in the objective function, or to repair them using local search methods \citep{Kramer2010}.

The design space of a production system topology is often highly constrained. For example, there can be constraints on the number of instances of a specific type, or on which ports are allowed to be connected \citep{Paape2023}. The consequence is that the number of permutations of a chromosome is often many times larger than the number of feasible designs. For instance, the small-scale industrial case study discussed in Section~\ref{sec:case_studies} has a chromosome of length $1156$, resulting in $2^{1156}$ possible solutions, of which only $11520$ are feasible. As a result, discarding, penalizing, or repairing infeasible solutions is impractical, as finding any feasible solution in a reasonable time frame is not guaranteed.

\begin{figure}[b]
    \centering
    \includegraphics[width=0.9\linewidth]{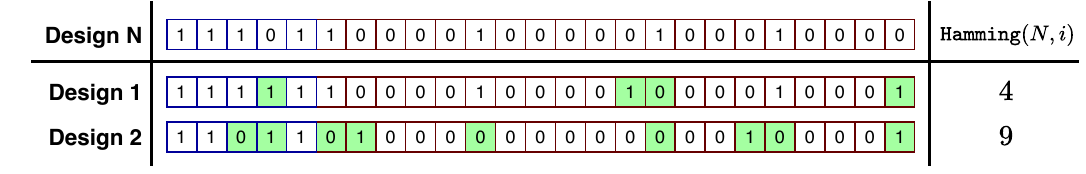}
    \caption{Calculation of the Hamming distances between design $N$ and respectively designs 1 and 2 from Figure~\ref{fig:design_space_example}. Highlighted in green are bits for which a design differs from design $N$. Design 1 has a higher similarity to design $N$, resulting in a higher chance for it to become the mutated offspring of design $N$.}
    \label{fig:hamming-distance}
\end{figure}

Instead, a similarity-based approach is taken comparable to the one proposed in \cite{Ismkhan2018}. The idea is to randomly select a parent solution from the population, and compare it to other known feasible solutions (not in the population). Offspring are then selected using the exponential rank-based selection algorithm based on their similarity to the parent; solutions with a higher similarity have a higher chance of being selected. Just as in \cite{Ismkhan2018}, similarity is calculated using the Hamming distance. For solutions $x$ and $y$, the Hamming distance is $\texttt{Hamming}(x,y) = \sum_{i}^{k} |x_{i} - y_{i}|$, with $k$ the number of bits in a chromosome. An example is shown in Figure~\ref{fig:hamming-distance}. 

The resulting algorithm for mutation is shown in Algorithm~\ref{alg:mutation}. The similarity-based approach has as its main benefit that there is no need for feasibility checks or repairs; only feasible solutions are considered. This is especially valuable when dealing with a highly constrained design space, in which the number of infeasible designs greatly outnumbers the number of feasible designs. However, this approach also has its drawbacks: calculating the similarity between designs becomes computationally expensive when the number of feasible designs is substantial. In such cases, it is more reasonable to consider only a subset of all feasible solutions. The set of potential children for which similarity is calculated is chosen randomly. A parameter $\delta$ is introduced which denotes the size of this set. In future work, it would be interesting to investigate if there are better methods for choosing the set of potential children. Another drawback to the similarity-based approach is that it requires all feasible designs to be known beforehand, which can be impractical when the number of feasible designs is especially large. 

\vspace{-.2cm}

\begin{algorithm}[tb]
\DontPrintSemicolon
\caption{The mutation algorithm}\label{alg:mutation}
\SetKwFunction{FnMutation}{Mutate}%
\Parameters{Mutation pressure $\alpha_m$, number of mutations $\gamma_m$, number of solutions to calculate the hamming distance for $\delta$}\;
\newcommand{\forcond}{$i=0$ \KwTo $\gamma_m$}
\Input{Population $X$}\;
\BlankLine
    Initialize mutated population $Y \leftarrow X$.\;
    \For{\forcond}{
        Randomly choose a random parent $p \in X$.\;
        Randomly choose the set of potential children $C$, a subset of the set of all feasible solutions $F$ of size $|C| = \text{min}(\delta, |F|)$.\;

        Calculate the similarity to $p$ of all solutions $c\in C$, with $\text{similarity}(c) = \texttt{Hamming}(p,c)$.\;

        Select a child $c \in C$ using rank-based selection based on similarity, with selection pressure $\alpha_m$.\;
        Add child $c$ to population $Y$.\;

    }
\BlankLine
\Output{Mutated population $Y$}\;
\end{algorithm}

\subsection{Recombination}
In recombination, also known as crossover, `genetic information' from two parent solutions is combined to produce offspring. It is inspired by the genetic recombination in biological reproduction. For recombination, the same similarity-based approach is taken as used for mutation. However, solutions are now compared to two randomly chosen parents and are selected using exponential rank-based selection based on the average similarity to these parents. The resulting algorithm is shown in Algorithm~\ref{alg:recombination}. 

\begin{algorithm}[htpb]
\DontPrintSemicolon
\caption{The recombination algorithm}\label{alg:recombination}
\SetKwFunction{FnRecombination}{Recombine}%

\Parameters{Recombination pressure $\alpha_r$, number of recombinations $\gamma_r$, number of solutions to calculate the hamming distance for $\delta$.}
\newcommand{\forcond}{$i=0$ \KwTo $\gamma_r$}\;
\Input{Population $X$}\;
\BlankLine
    Initialize recombined population $Y \leftarrow X$.\;
    \For{\forcond}{
        Randomly choose two parents $p_1, p_2 \in X$.\;
        Randomly choose the set of potential children $C$, a subset of the set of all feasible solutions $F$ of size $|C| = \text{min}(\delta,|F|).$\;

        
        Calculate the similarity to $p_1$ \& $p_2$ of all solutions $c \in C$, with $ \text{similarity}(c) = \frac{1}{2}\texttt{Hamming}(p_1,c) + \frac{1}{2}\texttt{Hamming}(p_2,c)$.\;
        
        Select a child $c \in C$ using rank-based selection based on similarity with selection pressure $\alpha_r$.\;
        
        Add child $c$ to population $Y$.\;
    }    
\BlankLine
\Output{Recombined population $Y$}\;
\end{algorithm}

\subsection{The \genetic algorithm}
The \genetic algorithm for topological optimization of production systems is based on the general form of the GA described in \cite{Back1993}, and is shown in Algorithm~\ref{alg:ga}. It utilizes the algorithms presented in the previous sections. The parameters of the algorithm and the values used in the case studies are shown in Table~\ref{tab:ga_parameters}. The selection pressure for recombination $\alpha_r$ was set higher than other selection pressures, with the reasoning that the random selection of the two parent solutions introduces enough randomization in itself.\\


\begin{algorithm}[H]
\DontPrintSemicolon
\caption{ The \geneticalgorithm for topological optimization of production systems using similarity-based mutation and recombination, discrete-event simulation-based evaluation, and rank-based selection.}\label{alg:ga}
\Parameters{population size $\beta$.}\;
\Input{Set of feasible solutions in the design space $F$.}
\BlankLine

Randomly choose an initial population $P \subseteq F$ of size $\beta$\label{line:ga_comptime1}.\;
\textbf{Evaluate} solutions in $P$ using DES; initialize set of evaluated solutions: $E \leftarrow P$.\;
\While{termination conditions are not met}{
    \textbf{Recombine} solutions in $P$ and add them to population $P'$.\label{line:ga_comptime2}\;
    \textbf{Mutate} solutions in $P'$  and add them to population $P''$.\label{line:ga_comptime3}\;
    \textbf{Evaluate} solutions in $P''$ using DES; add them to set of evaluated solutions $E$.\;
    \textbf{Select} next population $P \subseteq P''$ of size $\beta$ based on evaluated performance.\label{line:ga_comptime4}\;
    
    
}
\BlankLine
\Output{Evaluated solutions $E$.}
\end{algorithm}

\begin{table}[htpb]
    \caption{The parameters of the GA, and the values used in the case studies.}
    \label{tab:ga_parameters}
    \centering
    \begin{tabularx}{\linewidth}{|c|X|c|c|}
    \hline
 \textbf{Parameter}& \textbf{Description}&\textbf{Value} \\ \hline
         $\alpha_s$ &  Evolutionary pressure for selection& 1.3 \\ \hline
         $\alpha_m$ &  Evolutionary pressure for mutation& 1.3 \\ \hline
         $\alpha_r$ &  Evolutionary pressure for recombination& 2.0 \\ \hline
         $\beta$ &  Population size& 30 \\ \hline
         $\delta$ &  Maximum number of solutions to compare parent solutions to in mutation and recombination& 20000 \\ \hline
         $\gamma_m$ &  Number of offspring created with mutation& 30 \\ \hline
         $\gamma_r$&  Number of offspring created with recombination& 10\\ \hline
    \end{tabularx}
\end{table}

\subsection{Results}
The performance of the \genetic algorithm is analyzed using the case studies described in Section~\ref{sec:case_studies}. All optimization experiments were repeated for 30 runs. Each run terminated when the best design was identified, or after a maximum of 1000 iterations.


\subsubsection{Industrial case study}
\begin{figure}[bht]
    \centering
    \begin{subfigure}{0.48\textwidth}
        \includegraphics[width=0.8\textwidth]{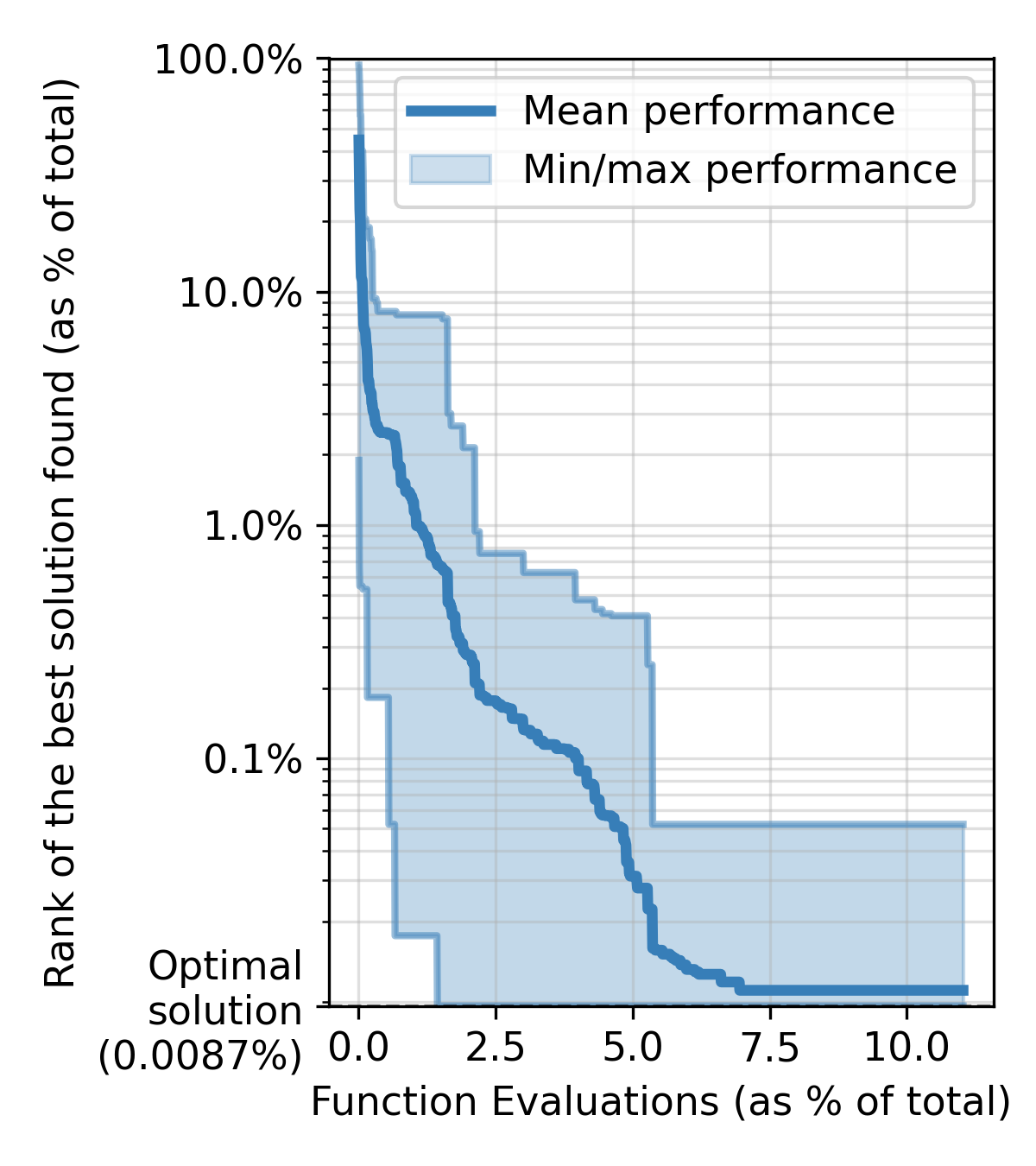}
        \caption{Small-scale case study (11520 designs)}
    \end{subfigure}
    \hfill
    \begin{subfigure}{0.48\textwidth}
        \includegraphics[width=0.8\textwidth]{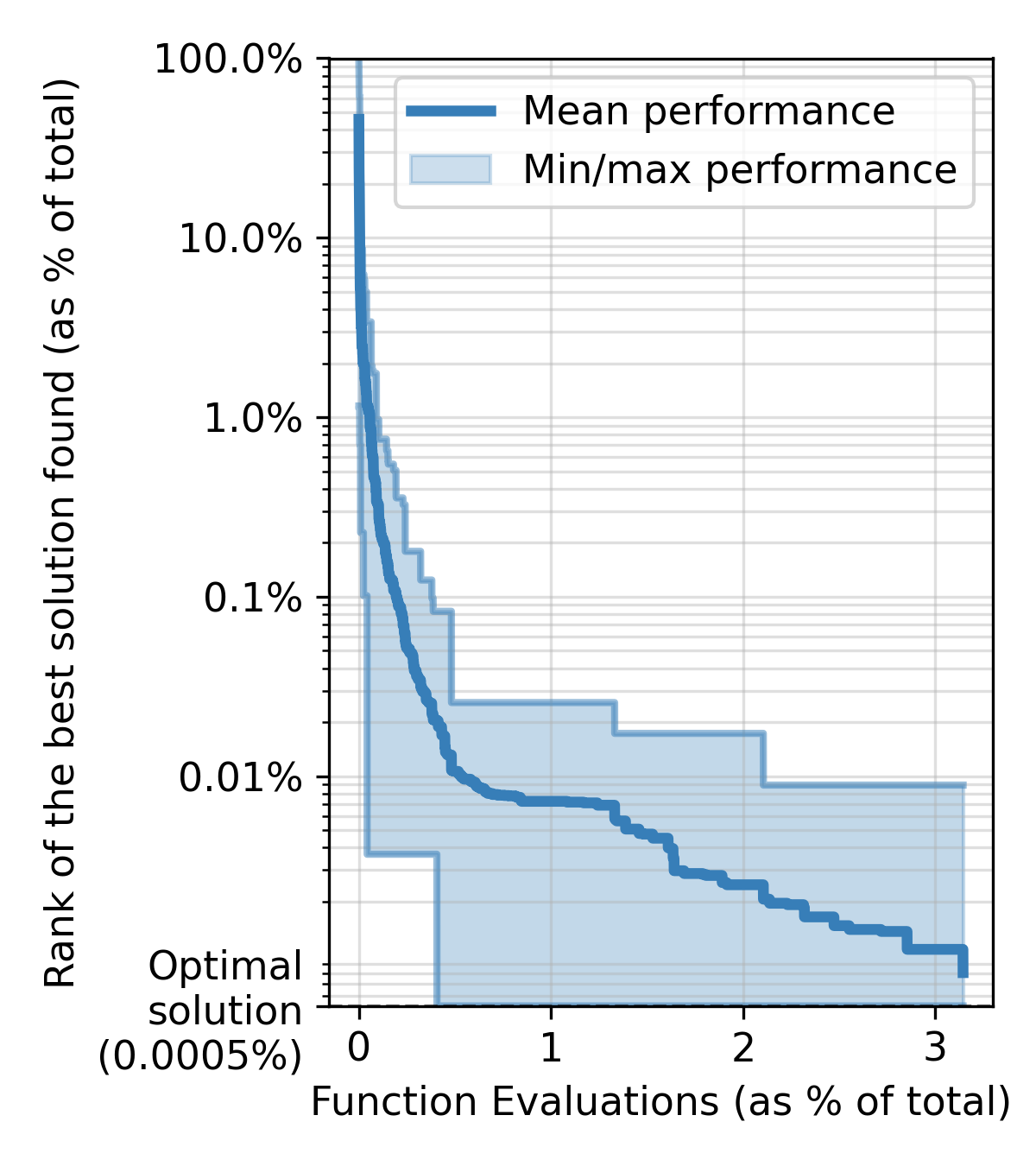}
        \caption{Large-scale case study (190464 designs)}
    \end{subfigure}
    \caption{The optimization performance of the GA for the industrial case studies.}
    \label{fig:opt_perf_ga}
\end{figure}

The \genetic algorithm was first evaluated on the small- and large-scale industrial poultry processing case studies described in Section~\ref{sec:poultry_case}. 
Figure~\ref{fig:opt_perf_ga} shows the optimization progress when using the GA for these two case studies. The horizontal axis shows the percentage of designs that have been evaluated through simulation. The vertical axis displays the ranking of the best design encountered so far as a percentage of all designs. The thick line shows the mean, and the colored area shows the minimum and maximum performance reached throughout the 30 runs of each experiment.

For both experiments, the optimal design was identified in 29 out of 30 optimization runs. For these runs, the mean number of function evaluations required to reach the optimal design (the size of output set $E$), was respectively 476 and 2609 evaluations. This corresponds to 4.13\% and 1.37\% of all total designs for the small- and large-scale case studies. 
This is a substantial improvement over the exhaustive search used in \cite{Paape2023c}, which on average required 50\% of all total designs to be evaluated. 
The average computation times of the algorithm were $4.0$ minutes and $27.3$ minutes for the small- and large-scale case studies respectively (excluding simulation time). 
Total simulation time was reduced from half a day to around $30$ minutes for the small-scale case study, and from two-and-a-half days to around $50$ minutes for the large-scale case study.

One observation was that the number of newly evaluated designs reduces after some iterations. At first, this would be close to 40 ($\gamma_m+\gamma_r$),  but after several iterations, it would be in the single digits, and sometimes no new designs are encountered at all. One reason is that as more designs are evaluated, the chance that offspring is already evaluated increases. Another reason is that the algorithm can get stuck selecting from the same group of designs because it searches based on similarity.

\subsubsection{Scalability case study}
\begin{figure}[htp]
    \centering
    \includegraphics[width=0.6\linewidth]{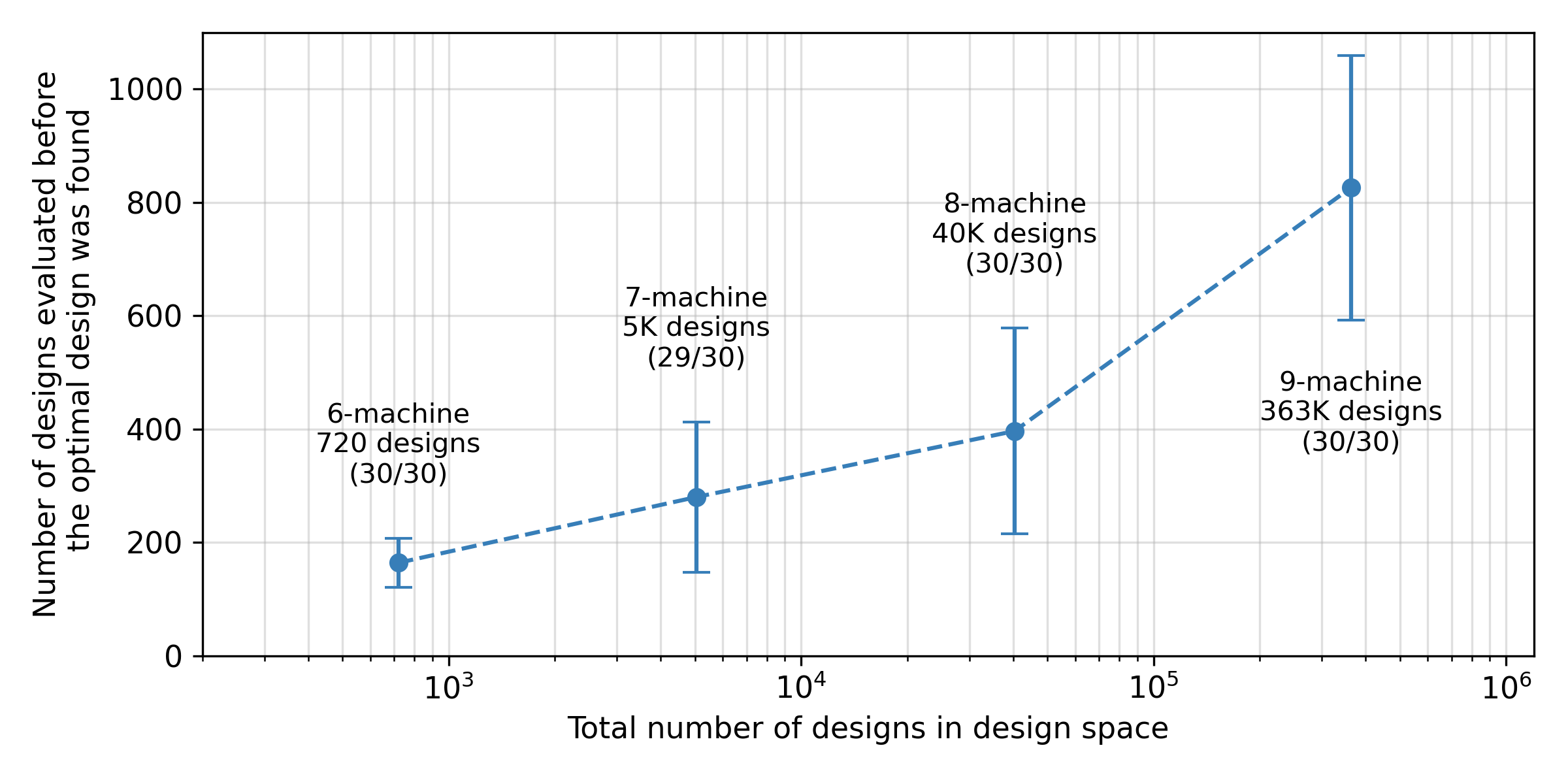}
    \caption{The scalability of the  genetic algorithm. The figure shows the mean and standard deviation of the number of designs that are evaluated before the optimal design is found in the 6-, 7-, 8- and 9-machine loop layout case studies. The number of designs in each case study, and the fraction of runs in which the optimal design was found, are displayed above each error bar.}
    \label{fig:ga_scalability}
\end{figure}

Next, the \genetic algorithm is evaluated on the loop layout case studies described in Section~\ref{sec:benchmark_case}, which serves as a benchmark for the scalability of the algorithm. The results are shown in Figure~\ref{fig:ga_scalability}. 
In all experiments, the optimal design was identified in all 30/30 runs, except for the 7-machine loop layout case study, where it was found in 29/30 runs. 
In all runs in which the optimal design was found, the mean number of evaluations required to find it were respectively 164, 280, 396, and 826 evaluations for the 6-, 7-, 8-, and 9-machine problems. This corresponds to respectively 22.8\%, 5.55\%, 0.983\%, and 0.228\% of all designs, indicating that the genetic algorithm scales well as the design space increases.


\section{The neural network-assisted genetic algorithm}\label{sec:nn-ga}
In itself, the \genetic algorithm can be used for optimization of production system topologies. In this section the \genetic algorithm of Section~\ref{sec:ga} is extended with a surrogate model, to reduce the need for computationally expensive discrete-event simulations. The surrogate model used is an artificial neural network that provides a computationally cheap approximation of the objective function. 

A neural network consists of layers of interconnected nodes organized into an input layer, hidden layers, and an output layer. An example of a neural network is shown in Figure~\ref{fig:feedforward_ann}. The input layer receives the initial data or `features', with each node representing one feature. The output layer produces the final prediction. Between the input and output layers, there are hidden layers where the network learns and extracts patterns from the input data. Each connection between nodes has a weight, representing the strength of the connection. These nodes apply an activation function to the weighted sum of their inputs. This activation step introduces non-linear characteristics to the network, allowing it to represent complex relationships within the data. The activation function used in this work is the  Rectified Linear Unit (ReLU) activation function \citep{Rasamoelina2020}.

The neural network is trained by feeding it with input data for which the output is known. The goal is to learn the relations between input and output in the underlying model. During training, the weights of the nodes are gradually adjusted over a number of `epochs' to minimize the difference between prediction and known output. This difference is expressed in a loss function. The loss function that is used depends on the chosen type of neural network. 

The training process is guided by an optimization algorithm that adjusts the network's weights to minimize the loss. 
The idea is that by learning from enough past observations, the network should be able to generalize well enough to make accurate predictions on unseen input data. A commonly used method for speeding up training, is to split the training set up into batches and to train on those, instead of training on all data at once. An extensive tutorial with more in-depth information on artificial neural networks can be found in \cite{Jain1996}.

Many different neural network architectures can be used for regression. In this work, three different types of architectures are compared. The first is the most commonly used neural network architecture; a feedforward neural network with fully-connected layers. The second type of architecture is a pairwise regression neural network. The third type of architecture is a Bayesian neural network (BNN). 
All three approaches utilize a frequently used optimization algorithm which has shown to be effective; the adaptive moment estimation algorithm, also known as the `Adam' algorithm \citep{Soydaner2020}. To improve generalization of the network a combination of popular techniques are utilized, which are: dropout \citep{Srivastava2014}, weight decay \citep{Krogh1991}, gradient clipping \citep{Zhang2019}, and early stopping \citep{Caruana2001}. 

In Section~\ref{sec:nn-architectures} the three types of neural architectures are explained. Section~\ref{sec:tuning} describes how the hyperparameters of the neural network are tuned.  

\subsection{Neural network architecture}\label{sec:nn-architectures}

\subsubsection{Feedforward neural network}
\begin{figure}[htpb]
    \centering
    \includegraphics[width=0.6\linewidth]{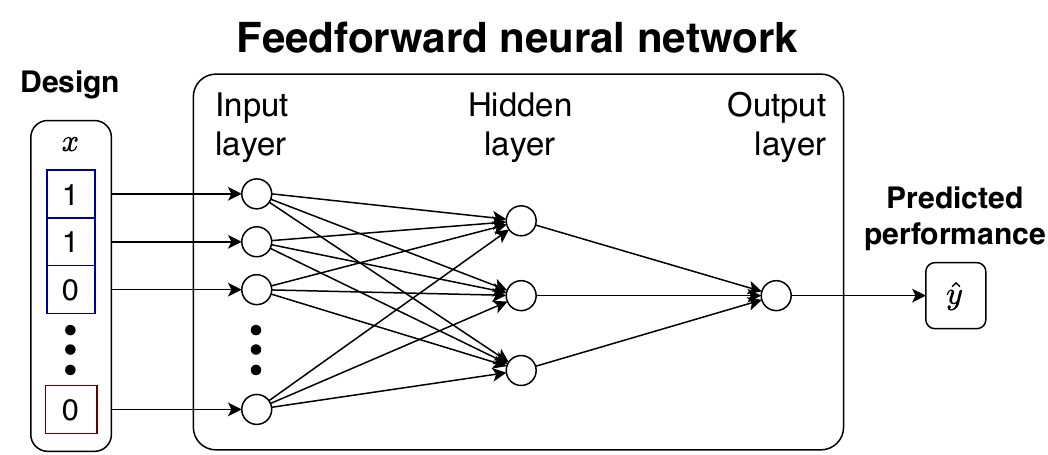}
    \caption{An example of a fully-connected feedforward neural network with one hidden layer. It takes as input the chromosome $x$ of a design, and as output gives a point estimate prediction $\hat{y}$ of the design's performance.}
    \label{fig:feedforward_ann}
\end{figure}

The first architecture is a fully-connected, feedforward neural network, which is generally seen as the `default' type of neural network. This type of architecture usually features one input layer, a number of hidden layers, and one output layer. Fully-connected means that the nodes of one layer are connected to all the nodes in the following layer. Feedforward means that information travels through the network in one direction, from input to output. Figure~\ref{fig:feedforward_ann} shows the architecture of a feedforward neural network with one hidden layer with three units.
For this architecture, the input and output are, respectively, the chromosome of a design, and its estimated performance. The network is trained on a set of solutions for which the performance is evaluated using discrete-event simulation. The loss function used is the mean squared error between target and prediction \citep{Wang2022}.


\subsubsection{Neural network with pairwise difference regression}
A pairwise neural network is a type of feedforward neural network, but instead of being trained to predict the performance of one design, it is trained to predict the performance difference between pairs of designs. The performance of a potential design is then predicted by comparing it with designs  for which the performance is already known. The idea is that for $n$ evaluated designs in the training set, there are $n^2$ pairs with which the model can train. By increasing the available training data, fewer evaluations are needed to effectively learn the underlying model. This approach is based on the work of \cite{Tynes2021}, in which pairwise regression is used to predict the properties of molecules using a feedforward neural network, and the work of \cite{Dushatskiy2019}, in which pairwise regression is used in combination with a convolutional neural network.

In this work, the architecture of the pairwise neural network is as follows. The input layer takes the chromosomes of two different designs as input; so for a chromosome of length $l$, the number of units of the input layer is $2\cdot l$. Next, are a number of fully-connected hidden layers. The output of the network is the predicted performance difference between the designs. This network is first trained on a training set for which the performance is known. This is done by iteratively sampling pairs of designs from this training set and learning the difference in performance. Again, the neural network is trained with the mean squared error as the loss function. 

\begin{figure}[btp]
    \centering
    \includegraphics[width=.95\linewidth]{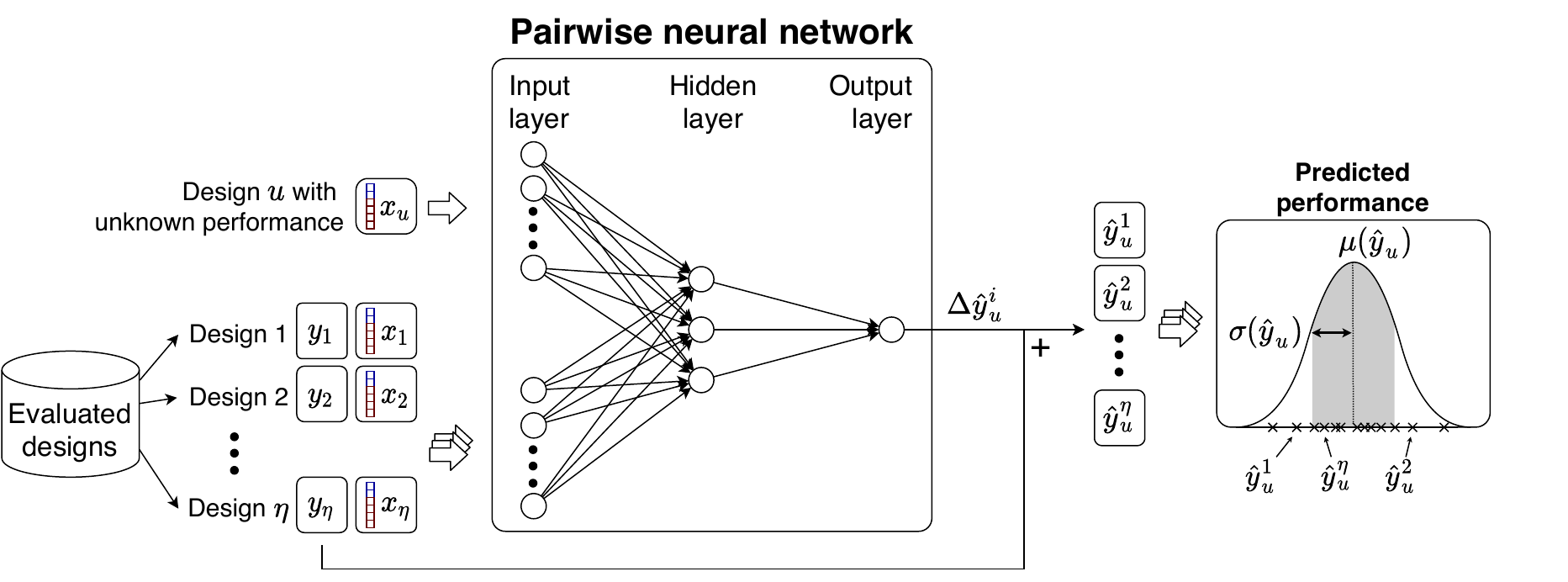}
    \caption{Example of a pairwise neural network. The network takes as input the chromosomes of two designs, and predicts performance difference $\Delta\hat{y}^i_u$. The performance of a design $u$ is predicted by comparing it with previously simulated designs $i=1, \ldots, \eta$, one at a time.     The known performances of these designs are used to calculate the individual predictions: $\hat{y}^i_u = y_i + \Delta\hat{y}^i_u$. These individual predictions are aggregated to determine the mean and standard deviation of the overall prediction of $\hat{y}_u$.
    %
%
    %
    %
    }
    \label{fig:pairwise_ann}
\end{figure}

After training, the neural network is used to make performance predictions. How this is done is shown in Figure~\ref{fig:pairwise_ann}. The input neural network again takes two designs as input; design $u$ with unknown performance, and a design $i$ from the set of evaluated designs. The predicted performance of design $u$ is the performance of design $i$ plus the predicted performance difference between the two. This process is repeated $\eta$ times, and the results are aggregated to predict the mean and standard deviation of the performance of design $u$.

\subsubsection{Bayesian neural network}
The third neural network architecture is the Bayesian neural network. A Bayesian neural network is a probabilistic model that incorporates uncertainty into its predictions. Instead of providing fixed point estimates, it estimates the distribution of possible outcomes. The advantages of BNN are that they are more robust to overfitting, and that they quantify the uncertainty in their predictions \citep{Liu2022}. Additionally, Our hypothesis is that the uncertainty estimation can be used to steer the \genetic algorithm toward solutions with high potential.

An extensive tutorial on BNN is given in \cite{Jospin2022a}. A BNN uses probability distributions for network weights. The initial beliefs about the distributions for these weights are called \textit{priors}, and they describe the initial assumptions on the model before any data is observed. As the BNN is trained and data is observed, these beliefs are updated to form the \textit{posteriors}, which are the updated weight distributions. Training a BNN requires a loss function that works with probability distributions instead of point estimates. The loss function for a BNN generally consists of two components, the negative log-likelihood (NLL) and the Kullback-Leibler (KL) divergence \citep{Jospin2022a}. The NLL measures how well the model explains the observed data; minimizing the NLL ensures that the network generates predictions that are consistent with the observed outcomes. The KL divergence measures the differences in the prior and posterior distributions; minimizing it ensures that the model stays close to the prior distributions, which helps prevent overfitting and improves generalization.

\begin{figure}[htpb]
    \centering
    \includegraphics[width=0.8\linewidth]{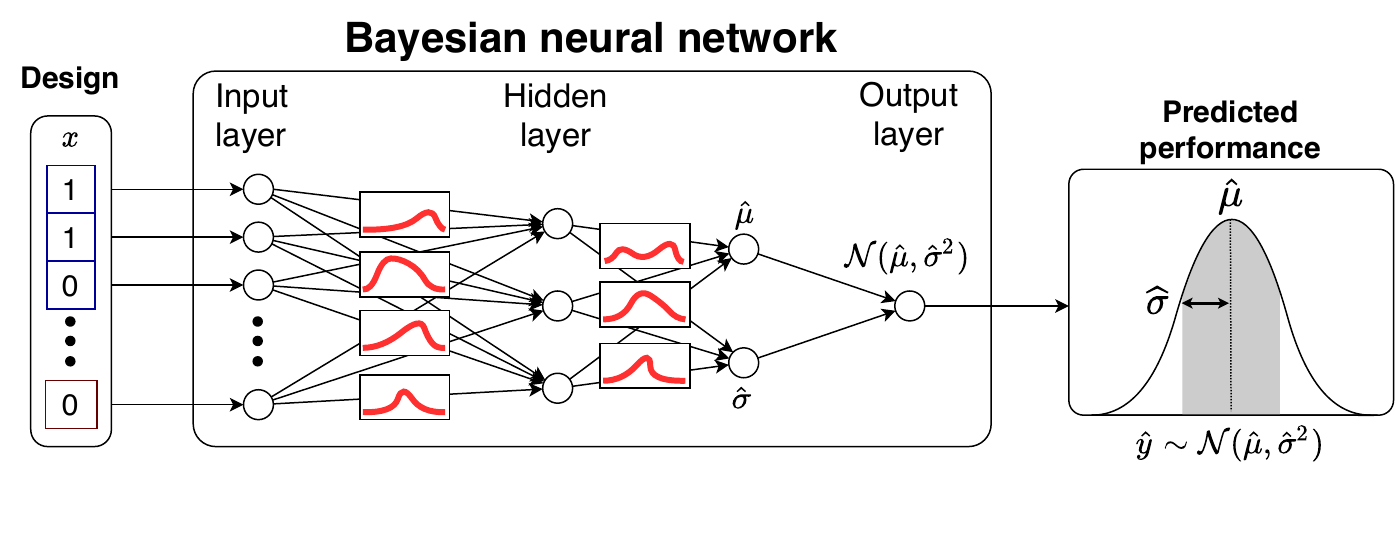}
    \caption{An example of a Bayesian neural network with one hidden layer. The BNN takes chromosome $x$ of a design as input, and as output returns the predicted normal distribution $\hat{y} \sim \mathcal{N}(\hat{\mu}, \hat{\sigma}^2)$. }
    \label{fig:bayesianANN}
\end{figure}

Figure~\ref{fig:bayesianANN} shows the architecture of the BNN, and how it makes predictions. Just like a normal feedforward ANN, it takes a chromosome as input, with one node in the input layer for each gene. What follows is a number of hidden layers. These are the layers that have probability distributions for network weights. These hidden layers are used to estimate the mean and standard deviation of the output distribution. 

\subsection{Tuning of the neural network}\label{sec:tuning}

One of the difficulties with using neural networks, is choosing the hyperparameters of the network. To create equal circumstances for all three types of neural networks, the hyperparameters are automatically tuned using random search \citep{Bergstra2012} for 50 trials. 
During tuning, different configurations of the neural network are evaluated. First, an initial set of solutions $T$ is selected for learning. The solutions in this set are evaluated using discrete-event simulation. The set is divided into a training set and a validation set using a 80/20 split. For each configuration of hyperparameters, the neural network is first trained on the training set. The model accuracy of this configuration is then determined based on the validation set. After all potential configurations have been analyzed, the configuration of hyperparameters which results in the highest accuracy is chosen. 

Some of the hyperparameters of the neural network are fixed, while others are tunable. For simplicity, all three neural network architectures feature only one hidden layer, which according to \cite{Macukow2016} is sufficient for a majority of problems. The number of hidden units in this layer is sampled from a range between $8$ to $128$ (with only powers of $2$ allowed). The activation function used in these units is the ReLU activation function. 
For training the Adam optimizer is used, with a maximum of 1000 epochs (although early stopping ensures that training is terminated when model accuracy no longer increases to prevent overfitting).
Another hyperparameter for training is the learning rate -- the rate at which the model learns from the  training data -- which during tuning is sampled exponentially between $10^{-4}$ and $10^{-2}$. Training is done in batches, and the batch size is allowed to be 8, 16, or 32. Other hyperparameters that are tuned are the dropout rate, weight decay, and gradient clipping norm. Table~\ref{tab:tuning_hyperparameters} gives an overview of the hyperparameters.


\begin{table}[htbp]
\caption{Fixed and tunable hyperparameters of the neural network.}
\label{tab:tuning_hyperparameters}
\begin{tabular}{|l|c|}
\hline
\textbf{Hyperpar.  (fixed)} & \textbf{Value} \\ \hline
Number of tuning trials   & 50     \\ \hline
Training/validation split & 80/20  \\ \hline
Number of hidden layers   & 1      \\ \hline
Optimizer                 & Adam   \\ \hline
Max epochs                & 1000   \\ \hline
Activation function       & ReLU  \\  \hline
\end{tabular}
\quad
\begin{tabular}{|l|c|c|}
\hline
 \textbf{Hyperpar. (tunable)} & \textbf{Min} & \textbf{Max} \\ \hline
Learning rate          & $10^{-4}$ & $10^{-2}$ \\ \hline
Batch size             & 8        & 32       \\ \hline
Number of hidden units & 8        & 128      \\ \hline
Dropout rate           & 0        & 0.5      \\ \hline
Weight decay           & $10^{-4}$ & $10^{-2}$ \\ \hline
Gradient clipping norm & 1.0      & 10.0   \\  \hline
\end{tabular}

\end{table}

\subsection{The neural network-assisted genetic algorithm}\label{sec:nn-ga-alg}

In the neural network-assisted algorithm, the network is used to determine which newly created offspring are most promising, and only those offspring are simulated. This allows the neural network-assisted genetic algorithm to explore the design space much more efficiently by reducing the dependency on computationally expensive simulations.

Algorithm~\ref{alg:nnga} shows how the neural network-assisted genetic algorithm works. At the start of the algorithm, a set of solutions $T$ is randomly chosen and evaluated through simulation. This set is used to tune and train neural network $\mathcal{M}$. After training the neural network, a subset of  $T$ is selected as the initial population. The population is recombined and mutated in the same way as in Algorithm~\ref{alg:ga}, but $\gamma_m$ and $\gamma_r$ are increased substantially so that more offspring are created. A prediction for the performance of these offspring is made using the neural network. Only the most promising solutions are then evaluated through simulation. 

How the most promising solutions are selected is slightly different for the three neural networks. The feedforward neural network makes a point-estimate for the performance of the solution, and solutions are selected using rank-based selection based on this prediction. When using the pairwise and Bayesian neural networks, the mean and standard deviation are estimated. These are used to calculate the probability that the predicted solution is better than the best solution found so far. By using this probability instead of the point-estimate, solutions with uncertain performance but high potential should have more chance of surviving. Solutions are selected using rank-based selection based on this probability. After simulating the most promising solutions, the best-performing solutions are selected using ranked-based selection. The neural network model is re-trained on the now larger dataset of evaluated solutions $E$ to improve the model accuracy further. This process is repeated till the terminating conditions are met.

The neural network-assisted GA uses different parameters than the unassisted GA. As said before, $\gamma_m$ and $\gamma_r$ are increased. Additionally, two new parameters are introduced. The first is the size of the initial dataset used for learning $\epsilon$. The second is $\phi$, the number of solutions that are evaluated with simulation in each iteration. The updated list of parameters and their values are shown in Table~\ref{tab:nnga_parameters}.

\begin{table}[btph]
    \caption{The parameters of the neural network-assisted GA, and the values used in the case studies.}
    \label{tab:nnga_parameters}
    \centering
    \begin{tabularx}{\linewidth}{|c|X|c|c|}
    \hline
 \textbf{Parameter}& \textbf{Description}&\textbf{Value} \\ \hline
         $\alpha_s$ &  Evolutionary pressure for selection& 1.3 \\ \hline
         $\alpha_m$ &  Evolutionary pressure for mutation& 1.3 \\ \hline
         $\alpha_r$ &  Evolutionary pressure for recombination& 2.0 \\ \hline
         $\beta$ &  Population size& 30 \\ \hline
         $\delta$ &  Maximum number of solutions to compare parent solutions to in mutation and recombination& 20000 \\ \hline
         $\gamma_m$ &  Number of offspring created with mutation& 250 \\ \hline
         $\gamma_r$&   Number of offspring created with recombination& 50\\ \hline
         $\epsilon$ & Size of the initial learning set & 100 \\ \hline
         $\phi$ & Number of solutions to evaluate each iteration & 40 \\ \hline 
    \end{tabularx}
\end{table}

\begin{algorithm}[phtb]
\DontPrintSemicolon
\caption{The neural network-assisted 
\genetic algorithm.}\label{alg:nnga}
\Parameters{size of the initial learning set $\epsilon$, number of solutions to evaluate $\phi$, population size $\beta$.}\\
\Input{Set of feasible solutions in the design space $F$.}
\BlankLine

Randomly choose initial set for learning $T \subseteq F$ of size $\epsilon$\label{line:nnga_nn_time1}.\;
\textbf{Evaluate} solutions in $T$ using DES. Initialize set of evaluated solutions: $E \leftarrow T$\label{line:nnga_evaluate_trainingset}.\;
\textbf{Tune} and \textbf{train} neural network $\mathcal{M}$ with the data collected on $T$.\label{line:nnga_nn_time2}\;
\textbf{Select} initial population $P \subseteq T$ of size $\beta$ based on evaluated performance.\label{line:nnga_comptime1}\;
\While{termination conditions are not met}{
    \textbf{Recombine} solutions in $P$ and add them to population $P'$\label{line:nnga_comptime2}.\;
    \textbf{Mutate} solutions in $P'$ and add them to population $P''$\label{line:nnga_comptime3}.\;
    Determine which solutions have not yet been evaluated: $P''_{\text{pred}} \leftarrow (P'' \setminus E)$.\label{line:nnga_nn_time3}\;
    \textbf{Predict} the performance of solutions in $P''_{\text{pred}}$ using neural network $\mathcal{M}$\label{line:nnga_nn_time4}.\;
    \textbf{Select} the set of solutions to evaluate $P''_{\text{eval}} \subseteq P''_{\text{pred}}$ of size $\phi$ using rank-based selection based on the predicted performance.\label{line:nnga_nn_time5}\;
    \textbf{Evaluate} solutions $P''_{\text{eval}}$ using DES, add them to set of evaluated solutions $E$.\;
    \textbf{Select} next population  $P \subseteq (P'' \cap E)$ of size $\beta$ from evaluated solutions in the population, based on evaluated performance.\label{line:nnga_comptime4}\;
    \textbf{Train} neural network $\mathcal{M}$ using the evaluated solutions $E$.\label{line:nnga_nn_time6}
}
\Output{Evaluated solutions $E$.}
\end{algorithm}

\subsection{Results}
In this section, the performance of the neural network-assisted genetic algorithm is analyzed. First, the performances of the three neural network architectures are compared using the small-scale industrial case study. To limit the number of optimization runs, one of the three architectures is selected to be used in the remainder of this work. The selection criteria are based on optimization performance and computation time.

\subsubsection{Comparison of neural network architectures as a surrogate model}
\begin{figure}[bth]
    \centering
    \begin{subfigure}{0.52\textwidth}
        \includegraphics[width=\textwidth]{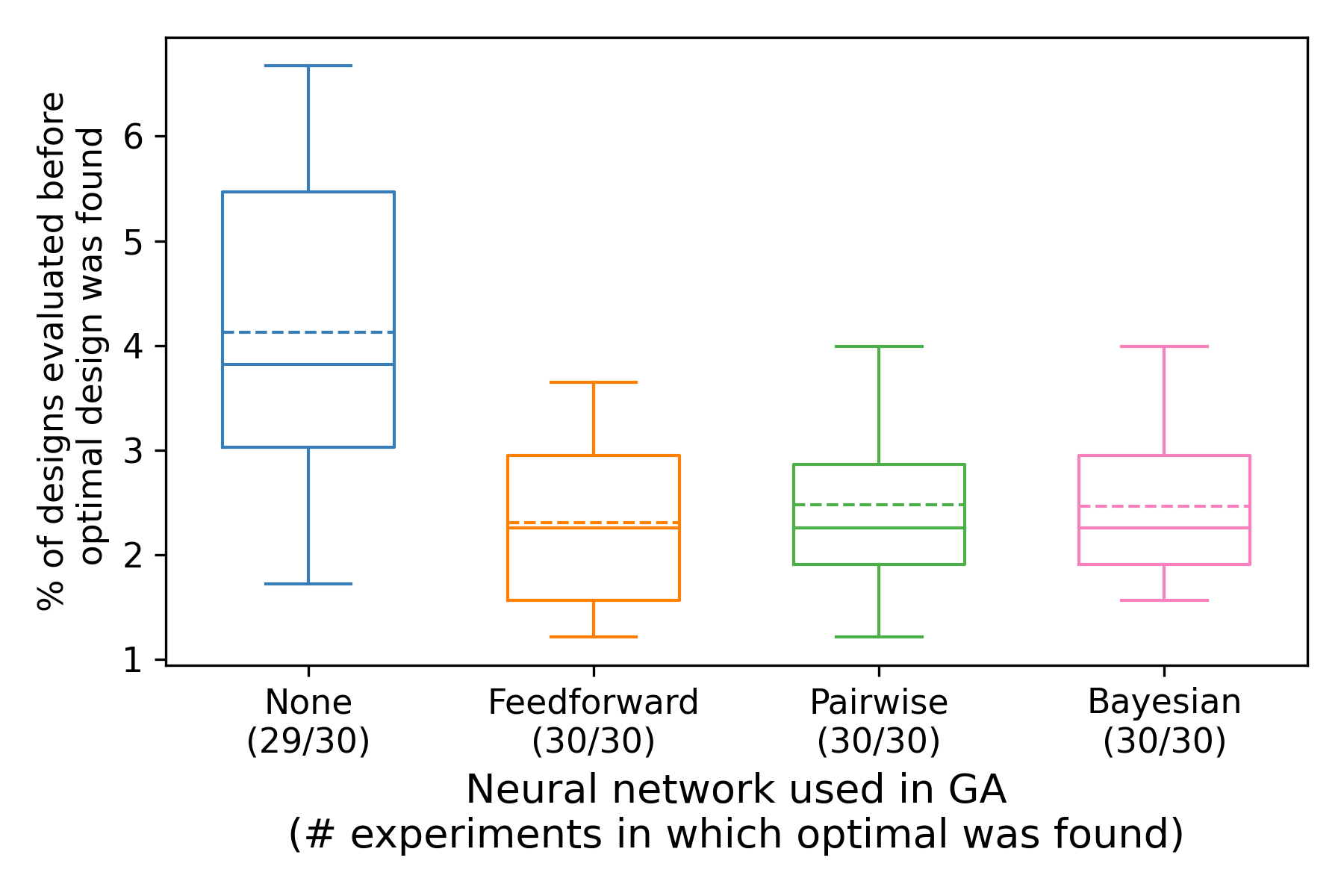}
        \captionsetup{format=hang}
        \caption{Optimization performance}
        \label{fig:nnga_architecture_comparison_perf}
    \end{subfigure}
    \hfill
    \begin{subfigure}{0.44\textwidth}
        \includegraphics[width=\textwidth]{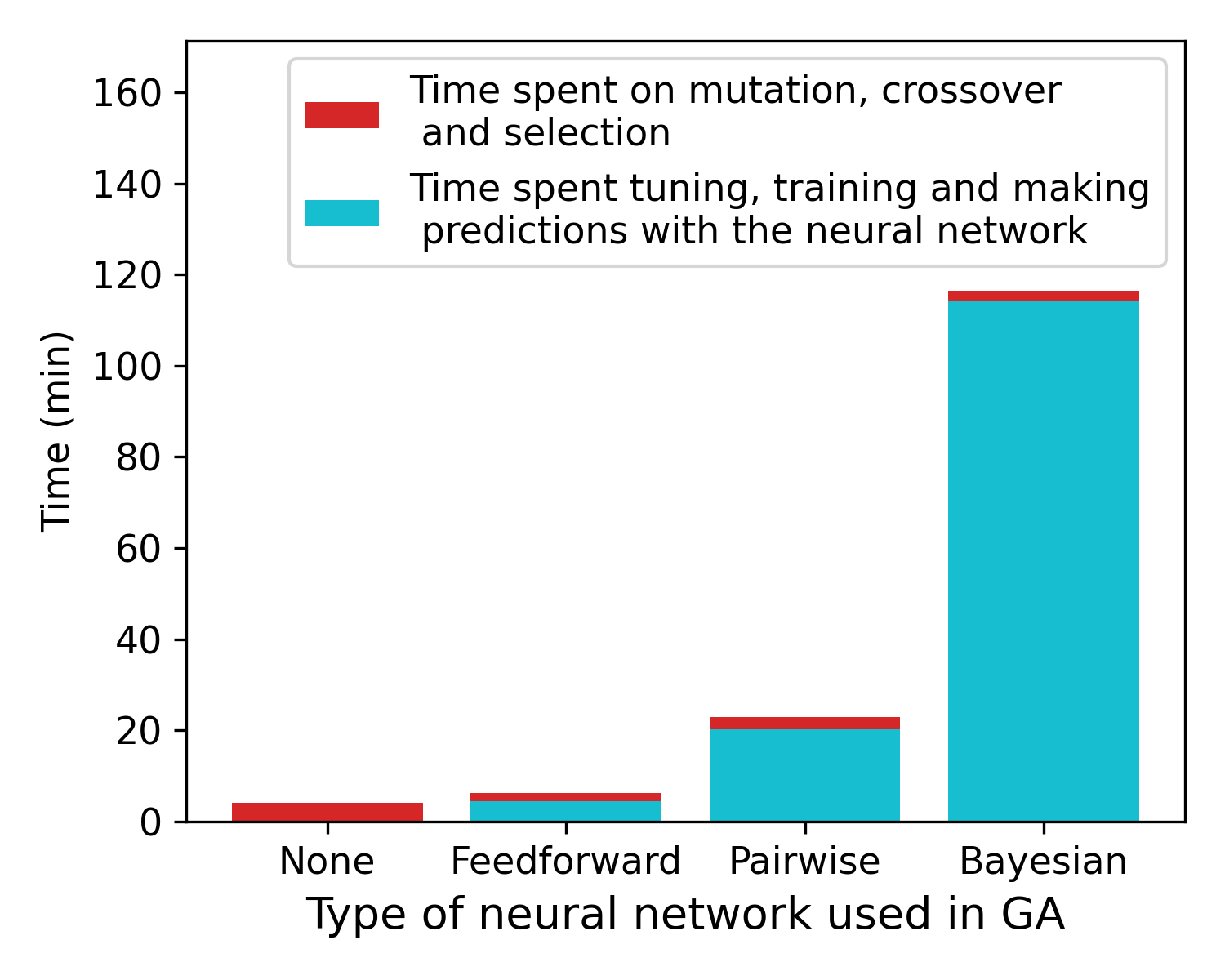}
        \captionsetup{format=hang}
        \caption{Computation time}
        \label{fig:nnga_architecture_comparison_time}
    \end{subfigure}
    \caption{A comparison of the (a) optimization performance, and (b) computation time, when using the unassisted GA with no surrogate model, or with either of the three neural network surrogate models, for the small-scale industrial case study.}     
    
    
    \label{fig:nnga_architecture_comparison}
\end{figure}

The neural network architecture that is used in the remainder of the paper is selected based on two metrics. The first and foremost is the optimization performance: is the optimal design found in every run of the optimization experiment, and if so, what is on average the \% of designs that need to be evaluated before it is encountered? The second metric is the computation time of the optimization algorithm. This metric acts as a tiebreaker in case of similar optimization performance. The priority is on optimization performance because the time it takes to simulate generally dominates the computation time of the algorithm, and reducing the number of evaluations directly reduces simulation time. 
Figure~\ref{fig:nnga_architecture_comparison} compares the optimization performance of the unassisted GA without a surrogate model, with the neural network-assisted GA using either of the three types of neural networks as a surrogate model.

Figure~\ref{fig:nnga_architecture_comparison_perf} shows a box plot depicting the optimization performance when using the four different approaches. For each approach, the results of the runs in which the optimal design is found are aggregated. The fraction of runs in which this was achieved is depicted under the respective boxes. The vertical axis shows the \% of designs that are evaluated before the optimal design is found. The solid horizontal lines of each `box' and `whiskers' depict the 2.5th, 25th, 50th, 75th, and 97.5th percentile for each approach, and the dashed lines depict the mean performance. The fraction of runs in which the optimal design was found is depicted below the respective boxes.

\begin{figure}[tbp]
    \centering
    \begin{subfigure}{0.32\textwidth}
        \includegraphics[width=\textwidth]{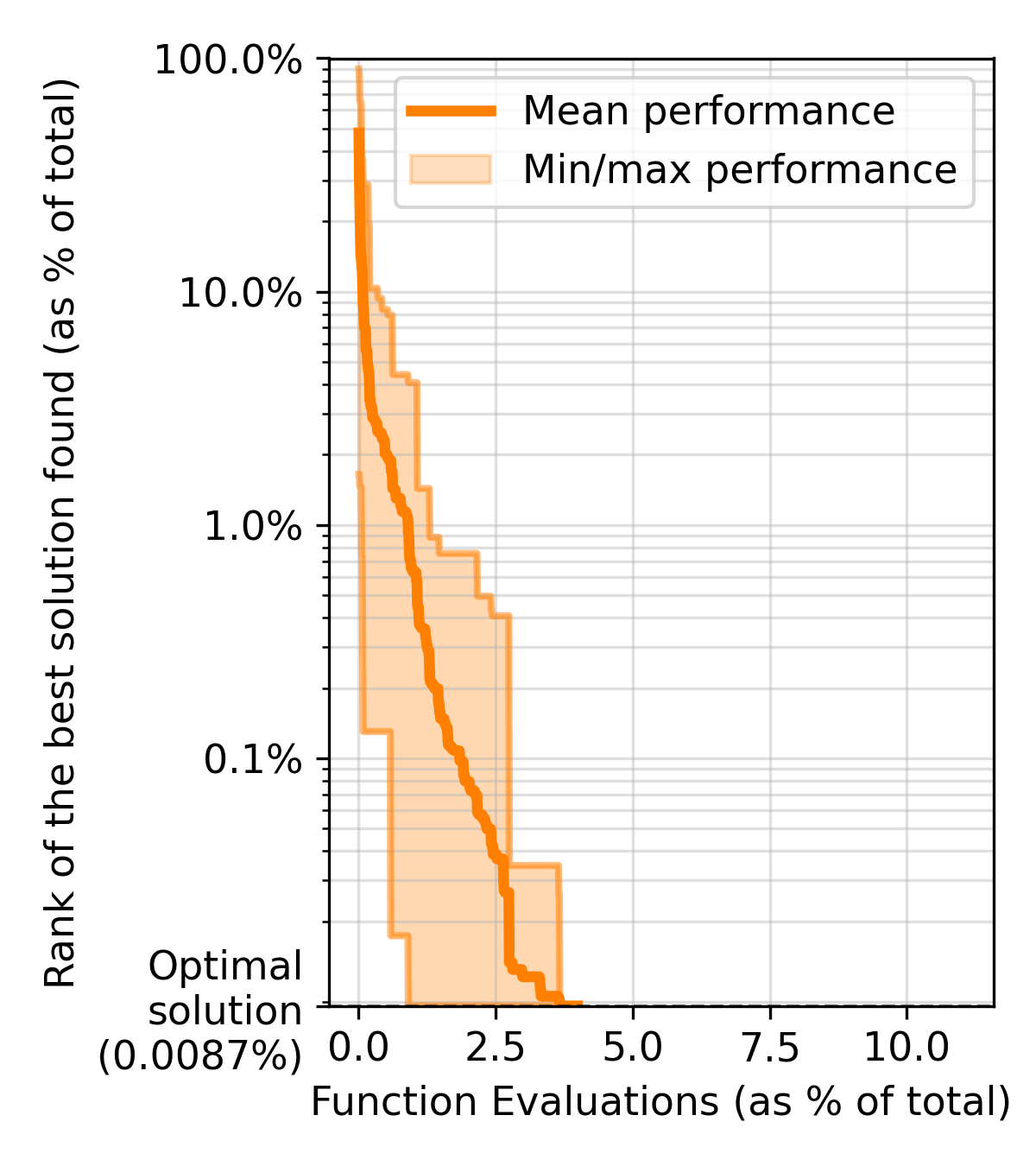}
        \captionsetup{format=hang}
        \caption{Feedforward ANN}
    \end{subfigure}
    \hfill
    \begin{subfigure}{0.32\textwidth}
        \includegraphics[width=\textwidth]{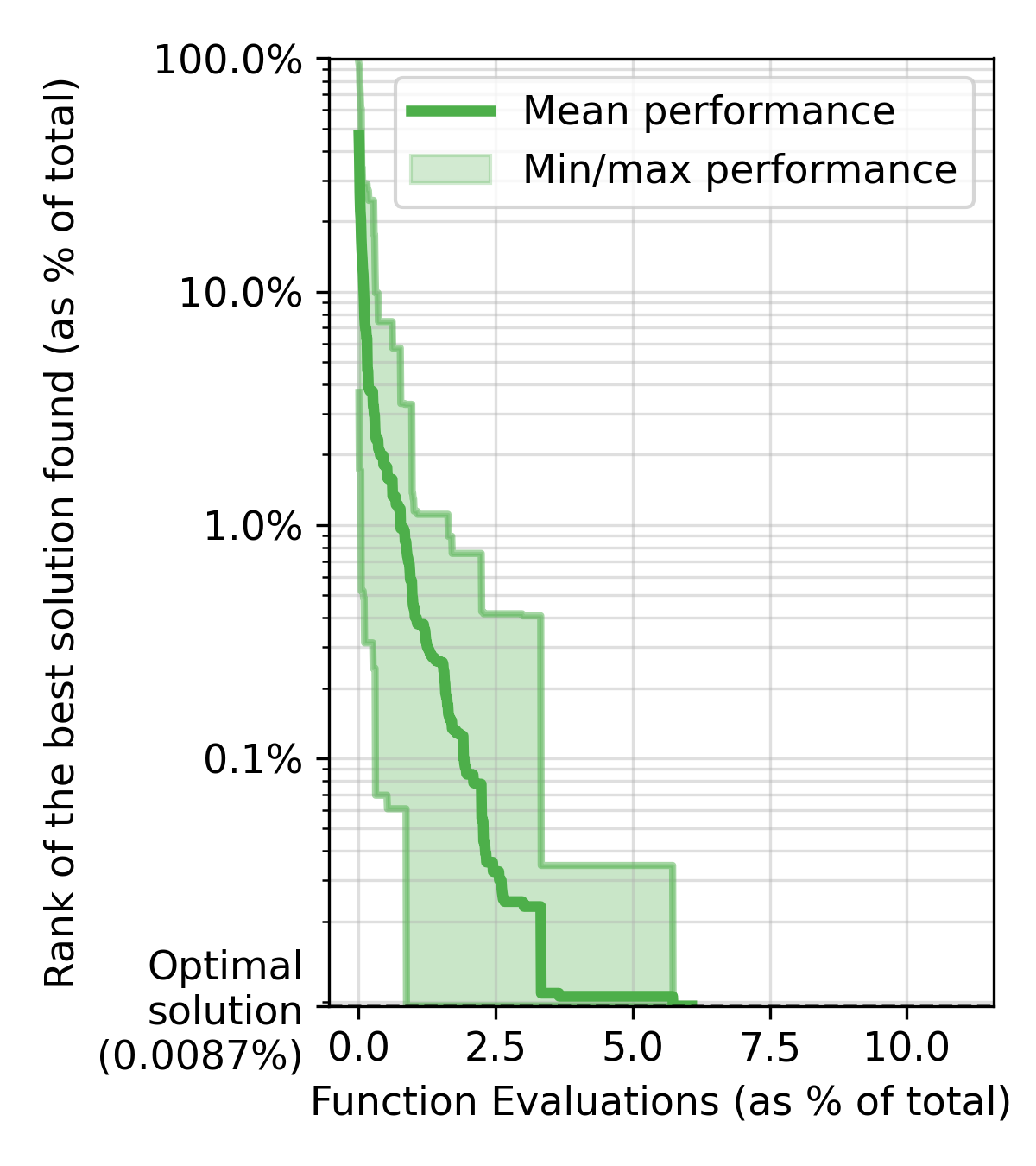}
        \captionsetup{format=hang}
        \caption{Pairwise ANN}
    \end{subfigure}
    \hfill
    \begin{subfigure}{0.32\textwidth}
        \includegraphics[width=\textwidth]{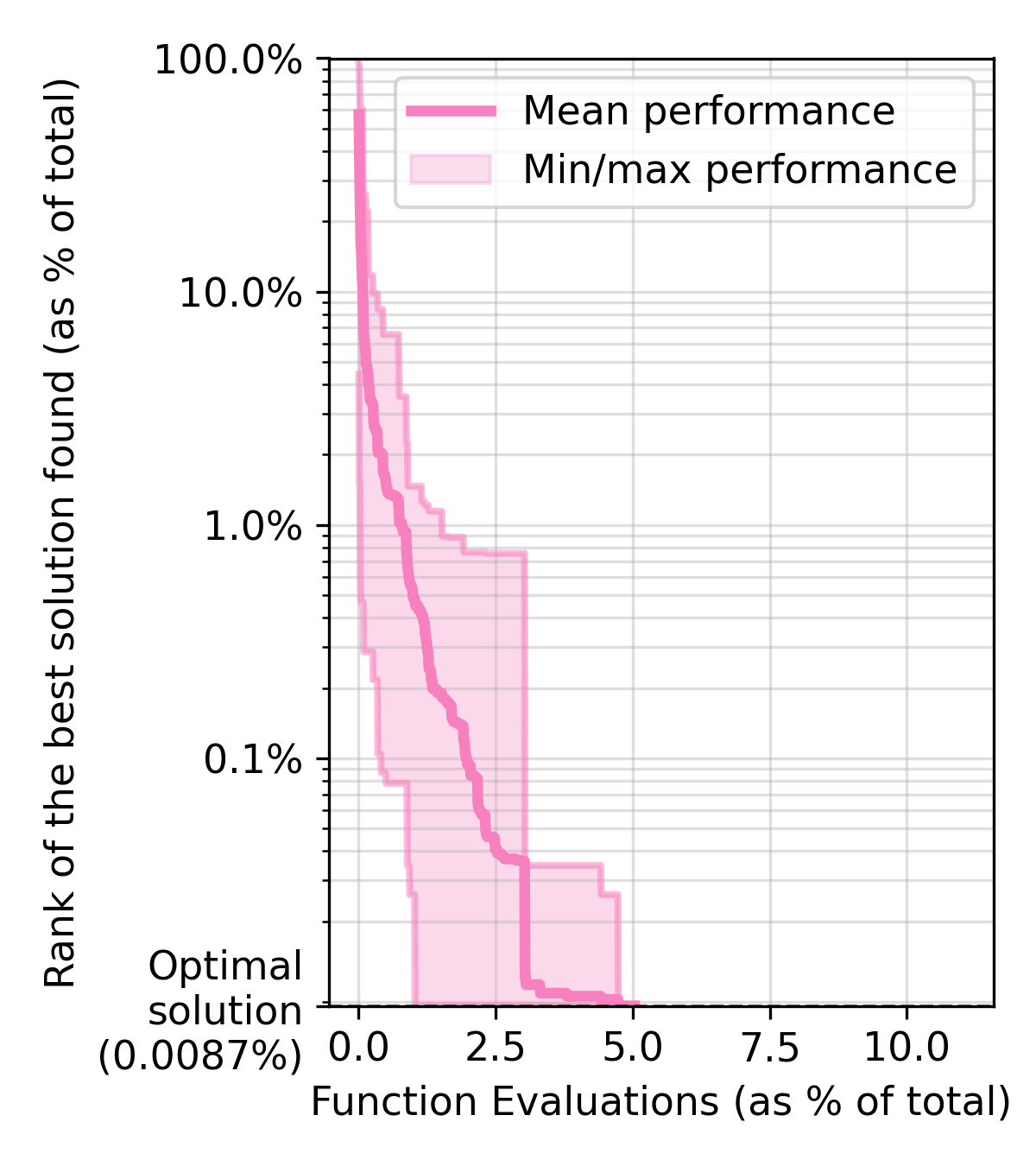}
        \captionsetup{format=hang}
        \caption{Bayesian ANN}
    \end{subfigure}
    \caption{The optimization progress throughout the use of the GA when using the three types of neural network as surrogate model in the small-scale industrial case study.
    }
    \label{fig:nnga_architecture_performance}
\end{figure}

In contrast with the unassisted GA (labeled as `None'), the optimal design was encountered in every one of the 30 runs when using a neural network surrogate model. For the unassisted GA on average, 576 designs were evaluated before the optimal design was found, which corresponds to 4.13\% of all designs. For the feedforward, pairwise, and Bayesian neural network-assisted GAs, it was respectively 265, 286, and 284 evaluations, corresponding to 2.30\%, 2.48\%, and 2.47\% of all designs. 
Figure~\ref{fig:nnga_architecture_performance} depicts the progress throughout the optimization process of the genetic algorithm when using the three types of neural networks as surrogate models. This figure shows that the progress throughout each of the three neural network-assisted GAs is also quite similar. One reason for the different architectures resulting in similar performance might be that a neural network can only learn so much if it is given a fraction of the designs in the design space; if trained well enough, all three neural network architectures will converge to a similar level of accuracy.

In Figure~\ref{fig:nnga_architecture_comparison_time} a comparison of the computation time of each variant is depicted. For the neural network-assisted algorithm, this is subdivided into the time spent on evolutionary optimization (lines~\ref{line:nnga_comptime1}, \ref{line:nnga_comptime2}, \ref{line:nnga_comptime3}, \ref{line:nnga_comptime4}), and the time spent on training and using the neural network surrogate model (lines~\ref{line:nnga_nn_time1}, \ref{line:nnga_nn_time2}, \ref{line:nnga_nn_time3}-\ref{line:nnga_nn_time5}, \ref{line:nnga_nn_time6}). The time it takes to simulate is excluded. It was observed that the majority of time spent on the neural network surrogate model is for tuning it. Especially for the pairwise and Bayesian neural networks, tuning took a significant amount of time. Only for the feedforward neural network the extra time spent on tuning and training the neural network weighs up against the reduction in simulation time due to less function evaluations. Because the optimization performance was almost identical for the three architectures, the feedforward neural network was selected for the remainder of this paper, as it has the lowest computation time. However, it must be said that due to the nature of neural networks, there are no guarantees on what performance will be achieved in future case studies. This might explain why pairwise regression has shown to be effective in \cite{Tynes2021} and \cite{Dushatskiy2019}, but does not show a performance difference in the case study analyzed here. From here onwards when referring to the neural network-assisted genetic algorithm in general, it is implied that the feedforward neural network is used.

\begin{figure}[tbp]
    \centering
    \begin{subfigure}{0.48\textwidth}
        \includegraphics[width=0.96\textwidth]{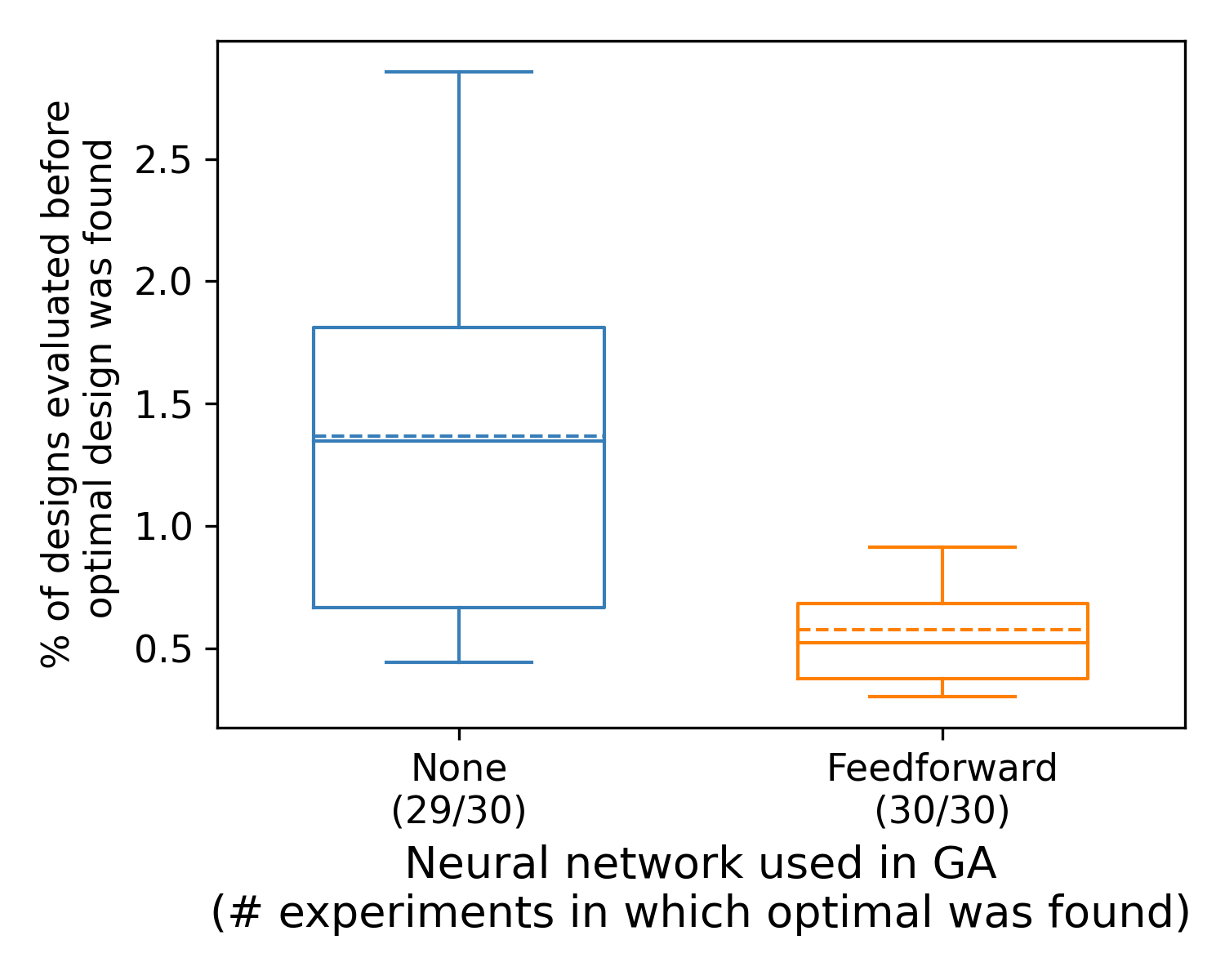}
        \captionsetup{format=hang}
        \caption{Optimization performance}
        \label{fig:nnga_large_scale_perf}
    \end{subfigure}
    \hfill
    \begin{subfigure}{0.48\textwidth}
        \includegraphics[width=0.96\textwidth]{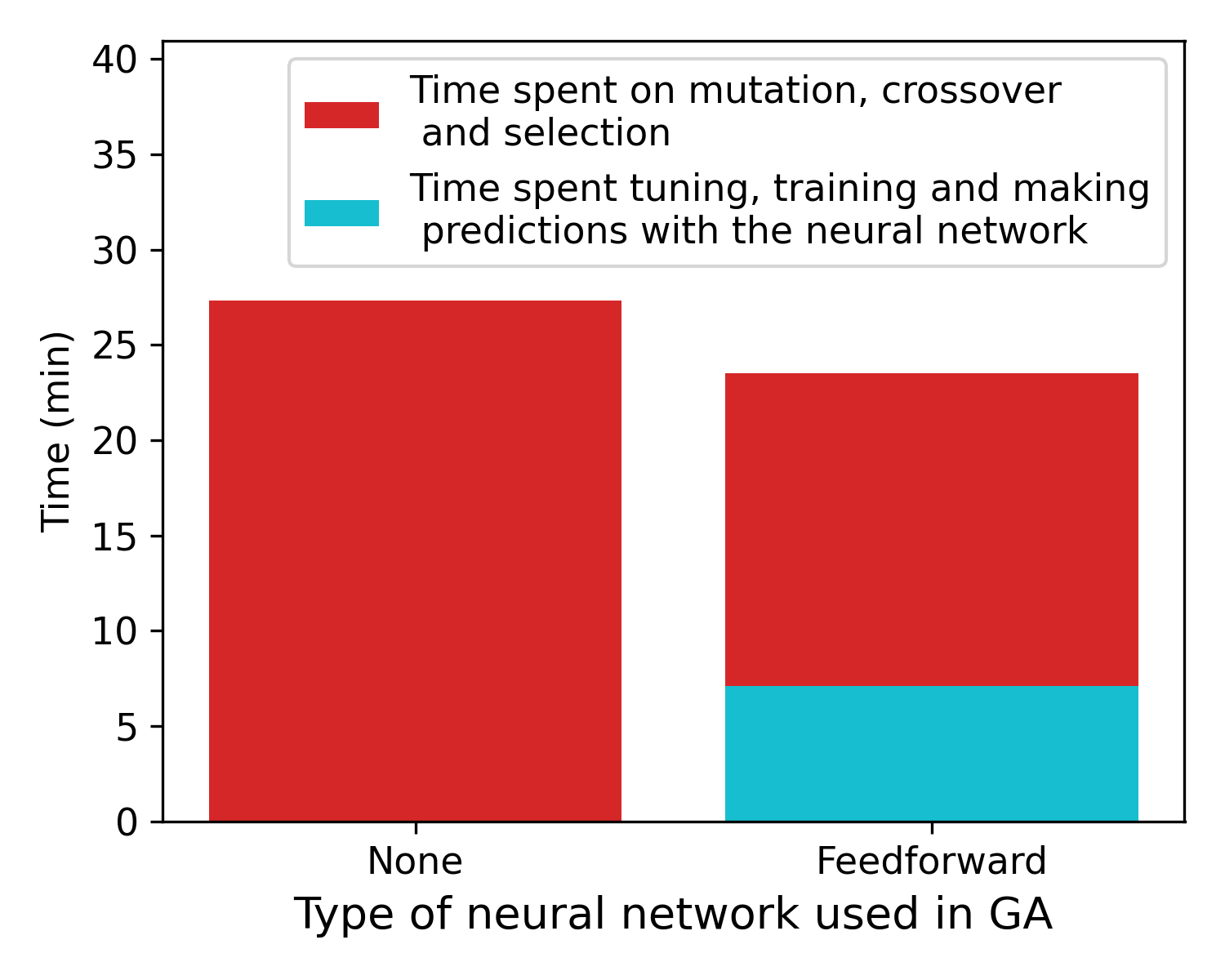}
        \captionsetup{format=hang}
        \caption{Computation time}
        \label{fig:nnga_large_scale_time}
    \end{subfigure}
    \caption{A comparison of the (a) optimization performance and (b) computation time for the unassisted GA, and the feedforward ANN-assisted GA, for the large-scale industrial case study.
    }
    \label{fig:nnga_large_scale}
\end{figure}

Figure~\ref{fig:nnga_large_scale} shows how the feedforward ANN-assisted GA compares to the unassisted GA for the large-scale industrial case study. For the feedforward ANN-assisted GA, the mean number of designs that were evaluated before the optimal design was found is 1009 evaluations, which corresponds to 0.53\% of all designs in the design space, compared to 1.37\% for the unassisted GA. The average computation time of the feedforward ANN-assisted GA is also lower than that of the unassisted GA, mostly due to fewer iterations being needed to find the optimum. Total simulation time was reduced from two-and-a-half days to around $20$ minutes for the large-scale case study.


\vspace{-.3cm}




\subsubsection{Scalability case study}
\begin{figure}[hbtp]
    \centering
    \includegraphics[width=0.8\linewidth]{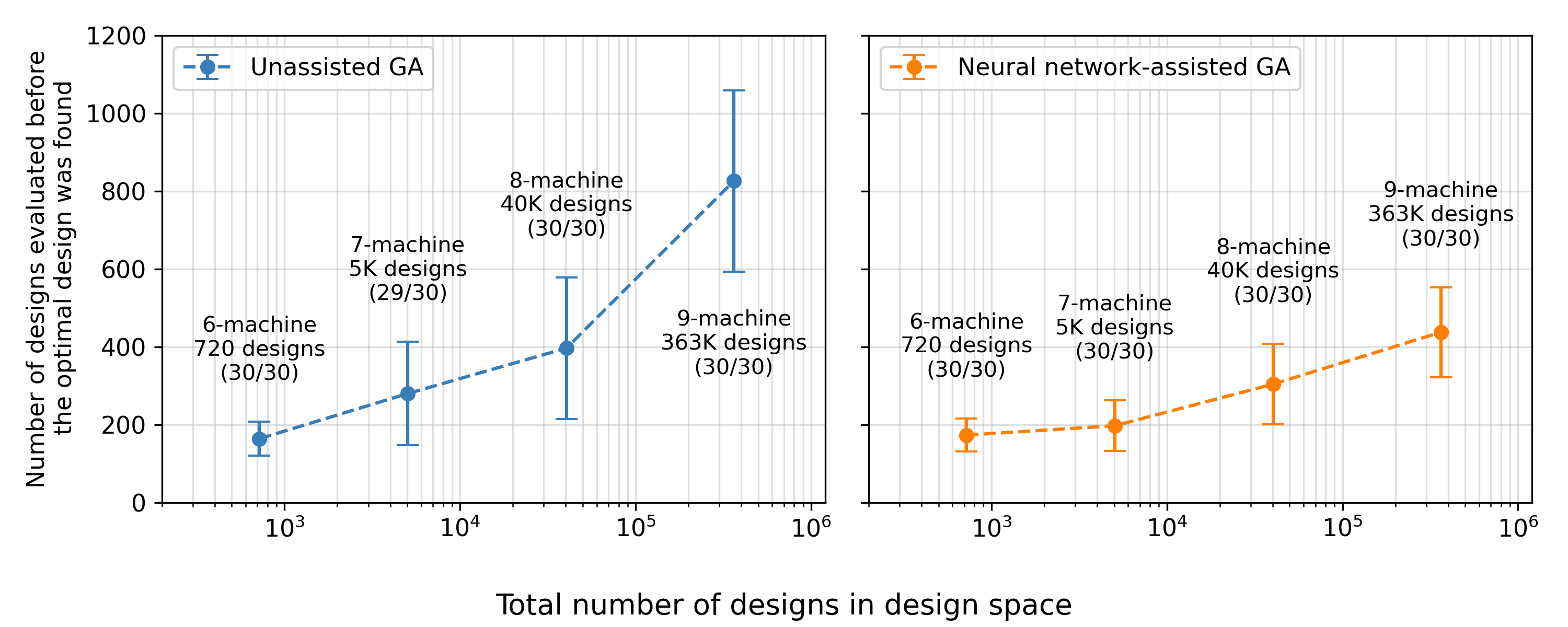}
    \caption{The scalability of the unassisted GA and the feedforward ANN-assisted GA. The figure shows the mean and standard deviation of the number of designs that are evaluated before the optimal design is found in the 6-, 7-, 8- and 9-machine loop layout case studies. The number of designs in each case study, and the fraction of runs in which the optimal design was found is displayed above each error bar.
    }
    \label{fig:nnga_scalability}
\end{figure}

The scalability of the feedforward ANN-assisted GA is evaluated on the loop layout case study described in Section~\ref{sec:benchmark_case}. 
A comparison of the scalability of the unassisted GA and feedforward ANN-assisted GA is shown in Figure~\ref{fig:nnga_scalability}. 
In all experiments, the optimal design was identified in all 30/30 runs. 
The mean number of evaluations required to find the optimal design were respectively 173, 197, 304, and 437 evaluations for the 6-, 7-, 8-, and 9-machine problems. This corresponds to respectively 24\%, 3.91\%, 0.754\%, and 0.121\% of all designs in the design space of each problem. In all but the 6-machine problem, the feedforward ANN-assisted GA outperforms the unassisted GA. The difference in optimization performance is the largest for the 9-machine problem, where the feedforward ANN-assisted GA requires almost half the number of evaluations (437 vs 826). This suggests that this approach has better scalability than the unassisted GA.

\section{Concluding remarks}\label{sec:conclusion}
In this work, a \geneticalgorithm is developed for topological optimization of production systems. Furthermore, an extension to the algorithm is presented for using a neural network to reduce the computational cost required for simulation. 

A method is proposed on how production systems topologies can be converted to a genetic representation. The general form of the genetic algorithm \citep{Back1993} is modified to handle production system topologies. The presented algorithm uses discrete-event simulation for fitness evaluation, rank-based selection, and similarity-based mutation and recombination operators. 

This genetic algorithm was further extended through a neural network surrogate model, to further reduce the computational cost required for simulation. Three types of neural network architectures were compared to determine which is most effective as a surrogate model in a genetic algorithm: a `default' feedforward neural network, a pairwise neural network, and a Bayesian neural network. In this specific comparison, the difference in optimization performance was negligible. However, the computation time of the feedforward neural network was substantially lower than that of the other approaches. An added benefit that advocates for the feedforward neural network is its simplicity compared to the other two types of architectures. 

Both the unassisted and neural network-assisted variants of the genetic algorithm were subjected to an industrial case study (with two different experiments) and a scalability case study (with four different experiments). Each experiment was repeated for 30 runs. The unassisted GA was able to find the optimal design in 177/180 runs, and the neural network-assisted GA was able to find it in all 180 runs. Both variants significantly reduce the number of evaluations required when compared to an exhaustive search. Furthermore, the scalability study showed that both variants scale well with the size of the design space; a 500-fold increase in the number of designs (from 720 to 362880) resulted in only a 5-fold increase in the number of evaluations for the unassisted GA, and a 2.5-fold increase for the neural network-assisted GA.

For future work, other approaches for mutation and recombination could be investigated. The current approach becomes less effective when the number of feasible designs increases, because it will take more and more time to calculate the similarity between all  designs. One idea could be to first subdivide the set of feasible designs into groups using a clustering algorithm. Another interesting research direction is to investigate whether control components can be included in the optimization of topologies, which would allow optimization of the production control architecture (e.g., different implementations of distributed production control). Finally, it would be interesting to identify whether other domains can benefit from the method presented in this paper. Although it has been developed for the production systems domain, there is no indication why it would not work for other domains in which solutions have similar topological structures, such as vehicle powertrain topology optimization \citep{Wijkniet2018}, or supply chain network optimization \citep{Bai2016}. After all, prior research has already shown that neural networks are also effective as a surrogate model in other areas of applications and for different types of simulation \citep{Yamada2022, Golzari2015, Esche2022}.

\bibliographystyle{tfcad.bst}
\bibliography{references}

\end{document}